\documentclass[journal]{IEEEtran}
\usepackage[pdftex]{graphicx}
\graphicspath{{./figures/}{./photos/}{./bio/}}
\DeclareGraphicsExtensions{.pdf,.jpg,.jpeg,.png}

\usepackage[utf8]{inputenc}
\usepackage[T1]{fontenc} 
\usepackage[cmintegrals]{newtxmath}
\usepackage{bm} 
\usepackage{amsmath}

\usepackage{hyperref}

\usepackage{array}
\usepackage{tabularx}
\usepackage[table]{xcolor}

\usepackage{bookmark}
\usepackage{booktabs}
\usepackage{multirow}

\usepackage{datetime}

\usepackage{threeparttable}

\newcommand{\ra}[1]{\renewcommand{\arraystretch}{#1}}

\ifCLASSINFOpdf
  \usepackage[pdftex]{graphicx}
\else
\fi
\ifCLASSOPTIONcompsoc
  \usepackage[caption=false,font=normalsize,labelfont=sf,textfont=sf]{subfig}
\else
  \usepackage[caption=false,font=footnotesize]{subfig}
\fi
\usepackage{pgfplots}
\usepackage{pgfplotstable}
\pgfplotsset{compat=1.7}
\usepackage{tikz}

\begin{document}
%
\title{Resource-aware Time Series Imaging Classification for Wireless Link Layer Anomalies}
%
%
%

\author{Bla\v{z} Bertalani\v{c},~\IEEEmembership{Graduate Student Member,~IEEE,}
        Marko Me\v{z}a,~\IEEEmembership{Senior member,~IEEE,}
        and~Carolina Fortuna\\%
    }

%
%

\markboth{Journal of \LaTeX\ Class Files,~Vol.~14, No.~8, August~2015}%
{Shell \MakeLowercase{\textit{et al.}}: Bare Demo of IEEEtran.cls for IEEE Journals}
%



\maketitle

\begin{abstract}
The number of end devices that use the last mile wireless connectivity is dramatically increasing with the rise of smart infrastructures and  require reliable functioning to support smooth and efficient business processes. To efficiently manage such massive wireless networks, more advanced and accurate network monitoring and malfunction detection solutions are required. In this paper, we perform a first time analysis of image-based representation techniques for wireless anomaly detection using recurrence plots and Gramian angular fields and propose a new deep learning architecture enabling accurate anomaly detection. 
We elaborate on the design considerations for developing a resource aware architecture and propose a new model using time-series to image transformation using recurrence plots. We show that the proposed model a) outperforms the one based on Grammian angular fields by up to 14 percentage points, b) outperforms classical ML models using dynamic time warping by up to 24 percentage points, c) outperforms or performs on par with mainstream architectures such as AlexNet and VGG11 while having <10 times their weights and up to $\approx$8\% of their computational complexity and d) outperforms the state of the art in the respective application area by up to 55 percentage points. Finally, we also explain on randomly chosen examples how the classifier takes decisions.

\end{abstract}
\IEEEpeerreviewmaketitle
\section{Introduction}
\IEEEPARstart{W}{ireless} networks represent the most convenient last mile connectivity solution and are used daily by billions of devices such as phones, tablets, laptops, desktops and increasingly smart devices forming the so-called Internet of Things~\cite{davies2020internet}. Traditionally, last mile connectivity issues were resolved by the owner of the access point be it an individual person, a cable or other operator in case of non-cellular operators, base station in the case of cellular only operators in a \textit{reactive} and \textit{mostly manual} manner. The reactivity is due to the fact that the entire model assumed that the customers would notice problems and take action by notifying the operators or resetting and reconfiguring equipment when possible. The manual mitigation is mostly due to the fact that the entire network, including the wired-to-wireless converter used static, \textit{pre-configured specialized} hardware equipment. 

\begin{figure*}[thb]
	\centering
	\includegraphics[width=\linewidth]{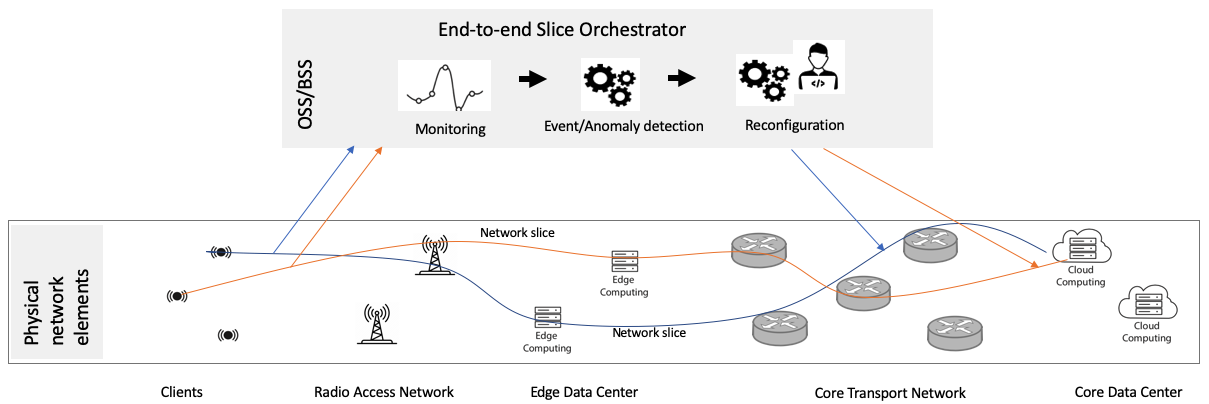}
	\caption{Example emerging wireless end-to-end slices and the role of automatic anomaly detection for the slice orchestrator.}
	\label{fig:bigpic}
\end{figure*}

With the increased digitization of society, including cities and infrastructures such as transportation and energy,  the number of end devices that use the last mile wireless connectivity is dramatically increasing and often require reliable functioning to support smooth and efficient business processes. Furthermore, the wired-to-wireless converters as well as the core transport networks are migrating to \textit{software controlled virtual functions residing on top of more general purpose hardware} leading to increasingly complex and expensive network operation. For instance, as shown in Figure \ref{fig:bigpic} and specified in the technical annex of the 3GPP 5G standard\footnote{3rd Generation Partnership Project, https://www.3gpp.org/}, an end-to-end network slice (i.e. virtual network service) can be created in a matter of minutes. The operational slices need to be continuously monitored throughout their lifetime and reconfigured as needed, possibly in a fully automatic manner, by the slice orchestrator in the OSS/BSS (operations/business support systems, respectively) of the network operators. 
The software and virtualization driven agility and complexity of the emerging networks requires faster configuration and reaction times. For instance, the automatic selection of a frequency band to be used at a given time and place should be suitably informed by events in the radio spectrum, automatically detected by automatic spectrum sensing, detection and classification systems \cite{gale2020automatic}, while the transmission parameters such as transmit power and channel number could be selected according to the perceived link quality \cite{9333616}. Finally, when a link outage or an abnormal network behaviour occurs, it should be proactively detected and fixed before it causes significant user dissatisfaction \cite{cerar2020anomaly} using network monitoring~\cite{silva2019m4dn} and malfunction detection~\cite{sheth2006mojo} solutions that automatically report relevant issues that can be mitigated without disrupting the business process.

An anomaly in general can be defined from a mathematical perspective as a rare event or an outlier of a distribution. Several machine learning techniques have been developed to detect such outliers and they typically use an unsupervised machine learning approach. For instance a typical example of such rare, unknown events that cannot be easily characterized a-priori is intrusion detection. Most such recent anomaly detections are performed on multivariate time series data \cite{thing2017ieee, ran2019semi, chen2018Wireless, Luo2018} using unsupervised deep learning algorithms in the form of autoencoder networks. 

An anomaly can also be defined from an application perspective as being an event that causes inconvenience to the respective application. The respective event does not necessarily need to be rare from a mathematical point of view, and its characterization is often known. For instance, Sheth~\textit{et al.}~\cite{sheth2006mojo} define and identify anomalies from the IEEE~802.11 physical layer perspective, namely, hidden terminal, capture effect, noise and signal strength variation anomalies, whereas Gupta~\textit{et al.}~\cite{gupta2007andes} define anomalies from multihop networking perspective, such as black hole, sink hole, selective forwarding and flooding. More recently, starting from observable symptoms of link measurements, namely the changes in the expected received signal, \cite{cerar2020anomaly} proposed four types of link layer anomalies, namely sudden degradation (SuddenD), sudden degradation with recovery (SuddenR), instantaneous degradation (InstaD) and slow degradation (SlowD). They were the  first to evaluate different time series representations such as Fast Fourier Transform (FFT), histogram, aggregates and compressed autoencoded in an attempt to develop a classifier for the defined anomalies. However, their approach only used classical, non deep learning methods. More recently, \cite{bertalanic_deep_nodate} investigated the same problem by training a deep learning model on raw time-series.

Inspired by the breakthroughs in image recognition~\cite{krizhevsky2012imagenet} and object recognition~\cite{zhao} that have been going on over the last decade, and attempts from various fields of science in formulating domain specific problems as image problems to benefit from these achievements\cite{singh, Sharma2019}, we also endeavour in the first  attempt to investigate image base transformations for supervised anomaly detection for wireless links as defined in \cite{cerar2020anomaly}. In this paper, we propose a new approach for anomaly classification in wireless networks based on a time-series to image transformation and deep learning. The contributions of this paper are:
\begin{itemize}
	\item We perform a first-time analysis of image based representation for wireless link layer anomaly detection. For this, we consider the four anomaly shapes defined in \cite{cerar2020anomaly}, the recently introduced Gramian angular fields \cite{wang2015encoding} (GAF) and recurrence plots \cite{eckmann1995recurrence}.
	\item  We elaborate on the design methodology to develop a new resource-aware deep neural network architecture for classification of the four types of anomalies. We propose a recurrence plot based model that outperforms the Gramian angular fields image-based models by up to 14 percentage points.
	\item  We also show the potential of the proposed model using recurrence plots to a) outperform classical ML models using dynamic time warping by up to 24 percentage points, b) outperform or performs on par with mainstream architectures such as AlexNet and VGG11 while having <10 times their weights and up to $\approx$8\% of their computational complexity and c) outperform the state of the art in the respective application area \cite{cerar2020anomaly,bertalanic_deep_nodate}, by up to 56 percentage points.
	\item  We show that the way the proposed model takes decision to classify instances can be explained.
\end{itemize}
This paper is organized as follows. We discuss related work in Section~\ref{sec:related}. Section~\ref{sec:problem_formulation} provides the formal problem statement, while Section~\ref{sec:TS_transform} elaborates on various time series transformations that can be used to generate image representations. Section \ref{sec:DL_models} introduces the proposed deep learning architecture, Section~\ref{sec:methodology} describes the relevant methodological and experimental details, while Section~\ref{sec:evaluation} provides thorough analyses of the results. Finally, Section~\ref{sec:conclusions} concludes the paper.
\section{Related work}
\label{sec:related}
 For detecting anomalies, models can be based on one of the four main categories of techniques: statistical, nearest neighbor, clustering or classification \cite{zhang}. Statistical models rely on the underlying data distributions which are normally not known for data with present anomalies. Techniques using nearest neighbor classification assume that normal patterns can be found in a dense neighborhood, while anomalies are far from it \cite{chandola2009anomaly}, but due to the nature of time series data this is not always the case. Clustering techniques work on a similar principle. Classification based techniques have shown superiority over other techniques in terms of learning patterns in the data, but their limitation is that they need labeled samples for training to learn how to discriminate between different patterns.
	
 Collecting a suitable training dataset for standard classification techniques is not always a feasible task. Since it is easier to assemble a dataset filled with normal data, one way to train a classifier for anomalies is with the use of the so called One-class classification techniques. These techniques are trained on a dataset filled with normal data patterns and then consider other patterns, that fall outside the learned boundary, as anomalies. Such techniques are particularly useful when it is not known how anomalous data looks like and are mostly interested in detecting if the data is anomalous or not. Binary classifiers mostly outperform One-class classifiers, but when the imbalance between normal and anomalous data is extreme One-class classifiers can outperform the binary ones~\cite{bellinger}. In our work we aim to detect specific shapes that are considered anomalies in the application area and use a suitable dataset thus focus on classification based techniques for developing binary and multiclass classifiers.

To support our contributions and put our work in perspective, we first analyze related work with respect to time series classification and then narrow down to works on image transformation for anomaly detection. 
\subsection{Time series classification in general}
 Unlike other type of data, time series contain ordered values and classification algorithms need to take this into account~\cite{langkvist2014review}. Begnall~\textit{et al.}~\cite{bagnall_great_2017} collected and investigated ways in which time series can be compared or classified into the correct class. Time series traces can be compared as a whole or by intervals, short patterns with shapelets can be detected, or a dictionary approach can be employed where the frequency of repetition of subseries is modeled and  classifiers are built based on the resulting histograms. A combination of all the previous methods can also be used as well as a customized model for each series and then measure similarities. With this in mind, Xing~\textit{et al.}~\cite{xing2010brief} divided time series classification (TCS) methods into three main categories: model-based, feature-based as used in this work, and distance-based methods. 
 
 Before the popularity of DL models, distance based classical ML models were mostly used for TS classification. For that purpose, researchers developed different distance metrics to improve the classification performance of the algorithm and one of the most successful ones is called Dynamic Time Warping (DTW)\cite{berndt1994using}. This metric is able to measure similarity, for example, between two TS traces that exhibit the same type of anomaly that occur at different time steps in each of the two traces.

L{\"a}ngkvist~\textit{et al.}~\cite{langkvist2014review} divided existing deep learning based models for TSC into two main categories: discriminative models and generative models. The two categories were further subdivided by Fawaz~\textit{et al.}~\cite{ismail_fawaz_deep_2019} into additional subgroups, namely models that use hand-engineering features and direct end-to-end models. End-to-end models are those where the input of a model is raw time series data, while feature engineering models take as input features created by hand through feature extraction, part of which are time series to image transformations. Eckmann~\textit{et al.}~\cite{eckmann1995recurrence} was probably the pioneer of time series to image transformations by introducing recurrence plots (RP). RPs were used mainly to see the data from a different perspective rather than to solve the classification problem of time series data. However, with DL and convolutional neural networks, RP became interesting for solving time series classification problems as noticed by Wang~\textit{et al.}~\cite{wang2015encoding}. In the same paper, they also introduced two other methods of time series to image transformations called Gramian Angular Fields (GAF) and Markov transition field. All three approaches have recently been used as an indirect method for classifying time series data using convolutional neural networks and deep learning networks. There are not many works on using time series to image transform for anomaly detection. All three transformations considered in \cite{wang2015encoding} were also used in solving regression problems \cite{bousbiat2020exploring}.
\subsection{Image transformation for anomaly detection}
The most recent usage of images for anomaly detection was presented by  Choi~\textit{et al.}~\cite{Choi9070362} that used Generative Adversarial Networks to transform multivariate time series into images and to detect and localize anomalies in signals from power-plants. Krummenacher~\textit{et al.}~\cite{Krummenacher8006280} did an interesting work on finding anomalies within wheels of train cart wagons by transforming sensor signal into Gramian Angular Field and using convolutional neural networks for detecting defects. One of the first applications of GAF for anomaly detection was attempted by Zhang~\textit{et al.}~\cite{zhang_automated_2019} for anomaly detection in EKG signals. Another interesting use of GAF and deep learning (DL) can be found in Xu\textit{et al.}~\cite{xu_human_2020} for recognizing human actions with signals received from wearable devices. This kind of classification is gaining momentum in the last few years, but as far as we know, no one used this approach for anomaly detection in wireless signals.

In recent years, some work has also been done on anomaly detection and time series classification using recurrence plots (RP). One of the first works on time series classification was by Silva~\textit{et al.}~\cite{silva_TS_RP_classification}, where they showed that transformation to RP improves classification performance on most of the datasets they used for testing. Hu~\textit{et al.}~\cite{hu_anomaly_2019} proposed a framework for anomaly detection in time series using RP transformation. Another anomaly detection work was done by~\cite{chen_cnn_rp} where they used RP transform and CNN on multisensor signals for real-time anomaly detection in flash butt welding process. In the network domain, an attempt was made by~\cite{kirichenko2019machine} to detect DDoS attacks using RP plots and convolutional neural networks. 
\section{Problem formulation}
\label{sec:problem_formulation}

\begin{figure}[thb]
	\centering
	\includegraphics[width=\linewidth]{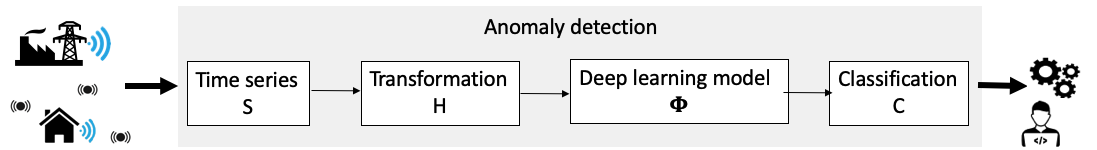}
	\caption{Wireless link anomaly detection steps.}
	\label{fig:problem_formulation}
\end{figure}

Starting from the example emerging wireless networks and the role of anomaly detection within depicted in Figure\ref{fig:bigpic}, we formulate the steps that enable automatic anomaly detection from time series. More specifically, we propose transforming the incoming time series data into images and then training a deep learning model to recognize anomalies in those respective images as depicted in Figure \ref{fig:problem_formulation} and further explained in this section. The model is then able to automatically classify  the images depending on whether they contain anomalies or not.
\subsection{Classification problem}
We formulate the anomaly detection problem as a classification problem in which given an input tensor H, there is a function $\Phi$ that maps the input to a set of target classes C as provided in Equation~\ref{mathref:classification}. 
\begin{equation}
	\label{mathref:classification}
	C=\Phi (H)
\end{equation}
The cardinality of the set C, also denoted as $|C|$, denotes the number of classes to be recognized. Without loss of generality, in the remainder of the paper we focus on two cases. In the first case, we consider a binary classification problem with $|C|=2$ where the set of target classes is $C=\{anomalous, normal\}$. In the second case, we start from the four types or anomalies introduced in~\cite{cerar2020anomaly} and consider a five class classification problem with $|C|=5$ where the set of target classes is $C=\{SuddenD, SuddenR, InstaD, SlowD, normal\}$. The binary classification problem aims to detect whether an incoming portion of a time-series contains or not one the four anomalies. The five class problem aims to detect the specific type of anomaly defined in the literature or a normal link.

\begin{figure*}[htbp]
	\centering
	\subfloat[Time-value perspective\label{fig:example:norecovery:ts}]{
		\includegraphics[width=.25\linewidth]{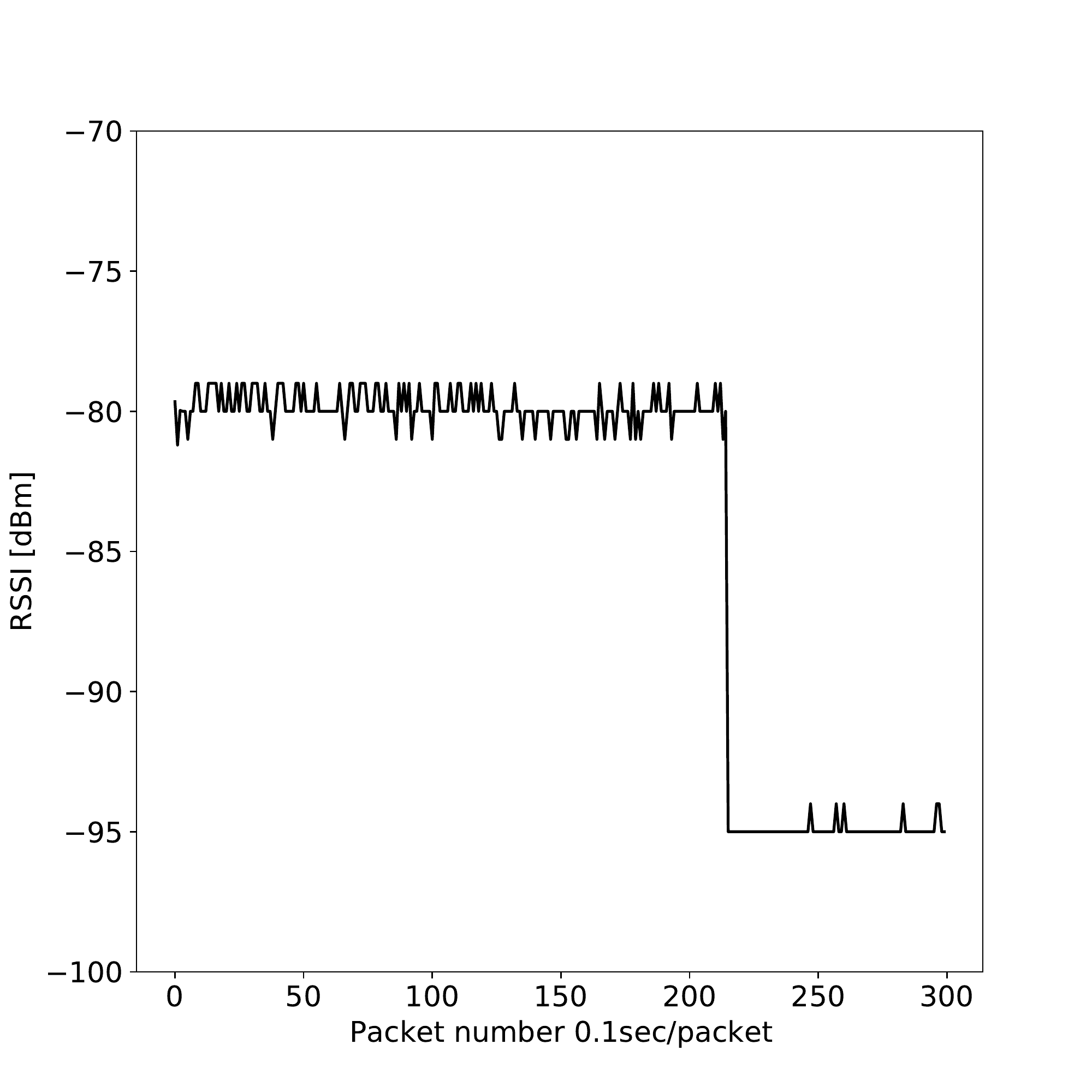}}%
	\subfloat[Recurrence plot transformation\label{fig:example:norecovery:rp}]{
		\includegraphics[width=.25\linewidth]{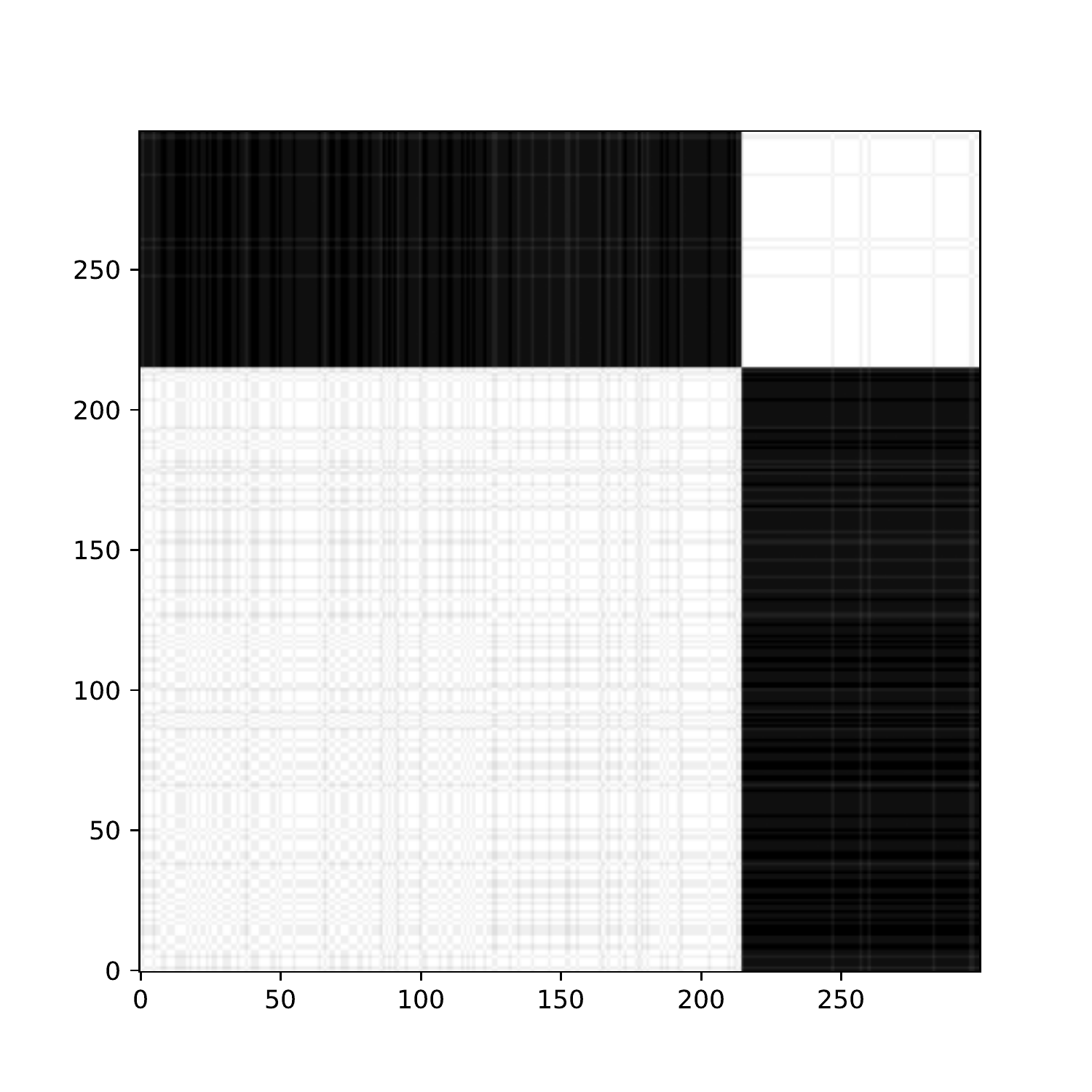}
	}%
	\subfloat[Gramian angular fields transformation\label{fig:example:norecovery:gaf}]{%
		\includegraphics[width=.5\linewidth]{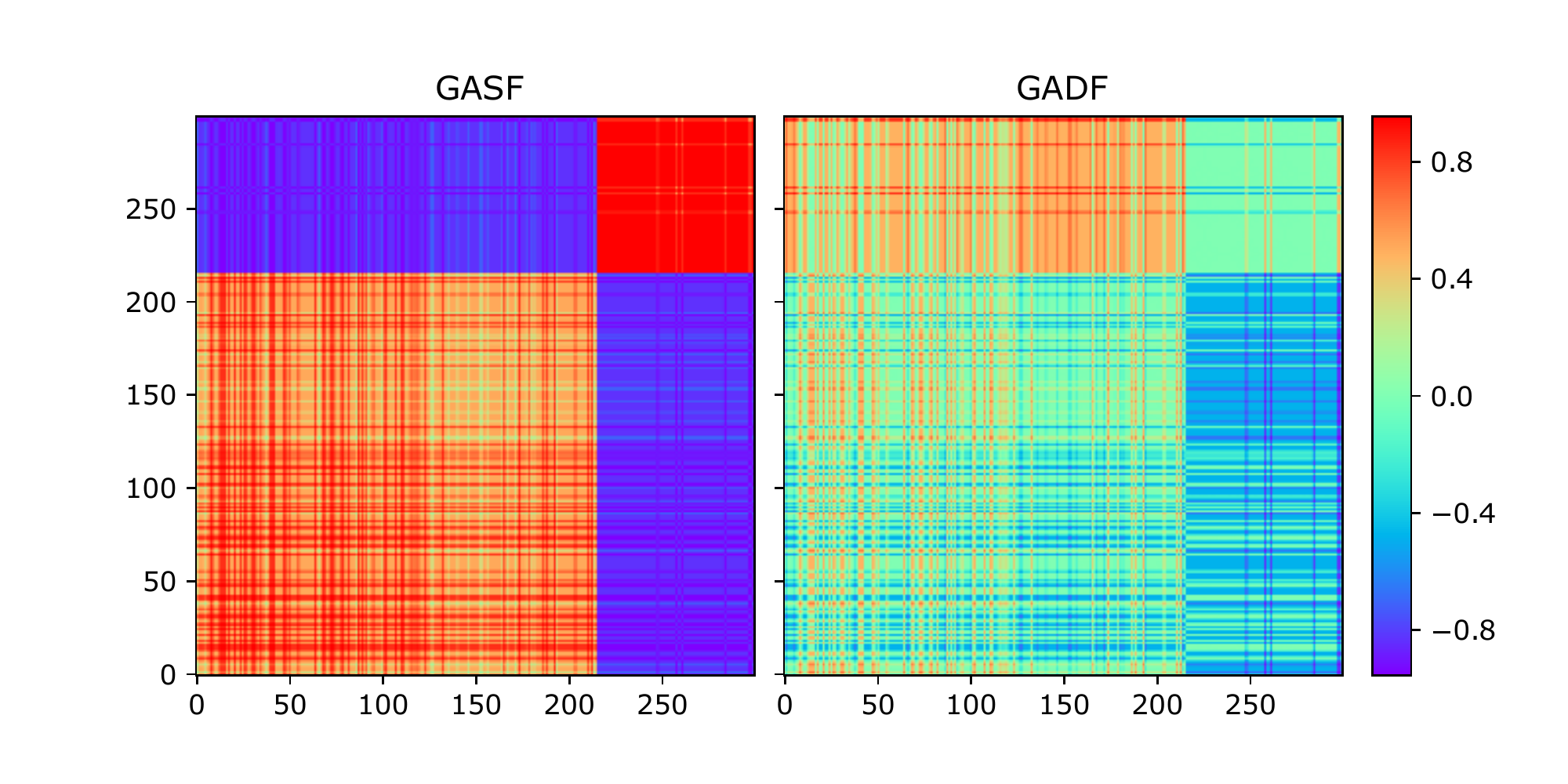}
	}%
	\caption{Distinct representations of the link layer RSSI measurement data for sudden degradation anomaly (SuddenD).}
	\label{fig:example:norecovery}
\end{figure*}

\begin{figure*}[htbp]
	\centering
	\subfloat[Time-value perspective\label{fig:example:step-recovery:ts}]{
		\includegraphics[width=.25\linewidth]{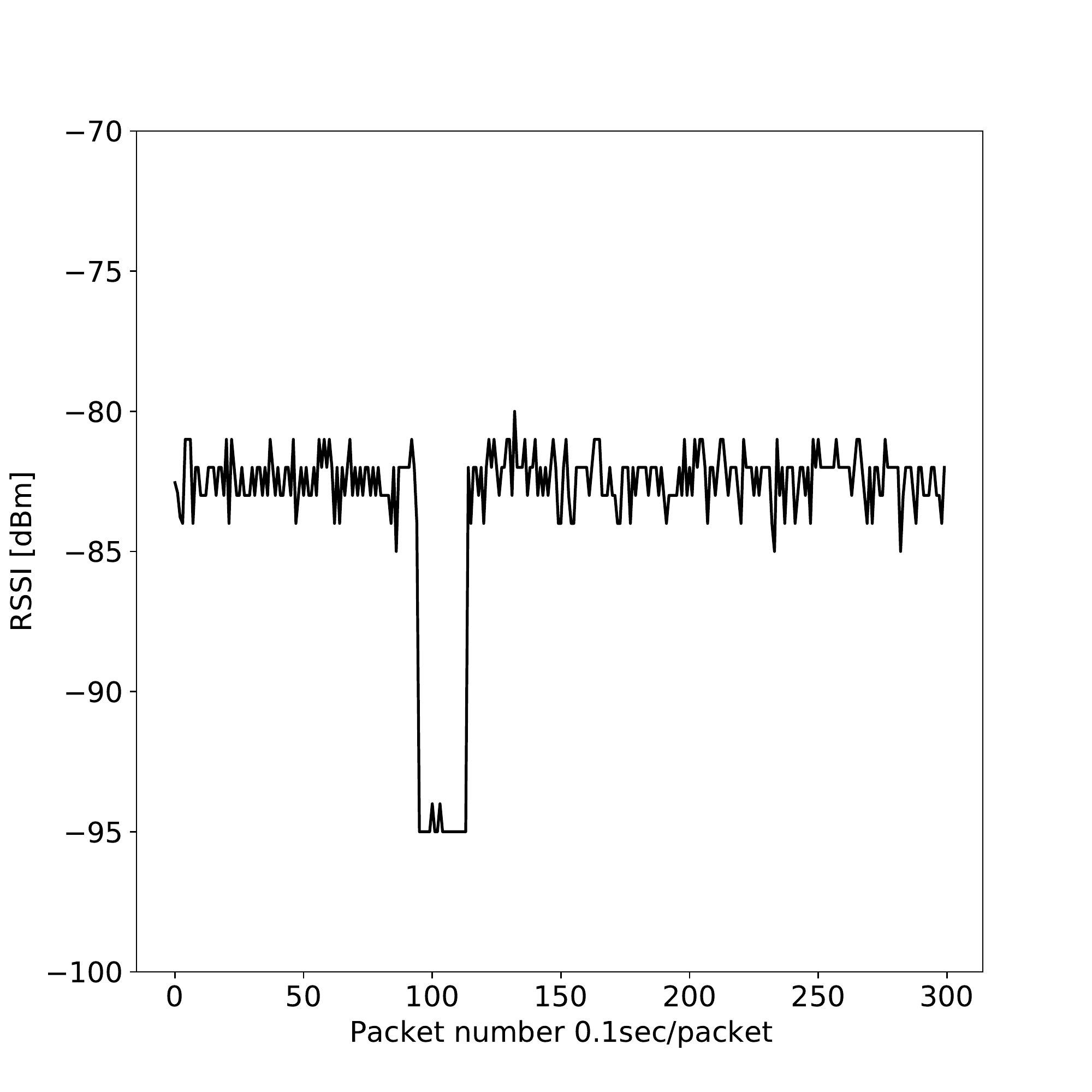}
	}%
	\subfloat[Recurrence plot transformation\label{fig:example:step-recovery:rp}]{
		\includegraphics[width=.25\linewidth]{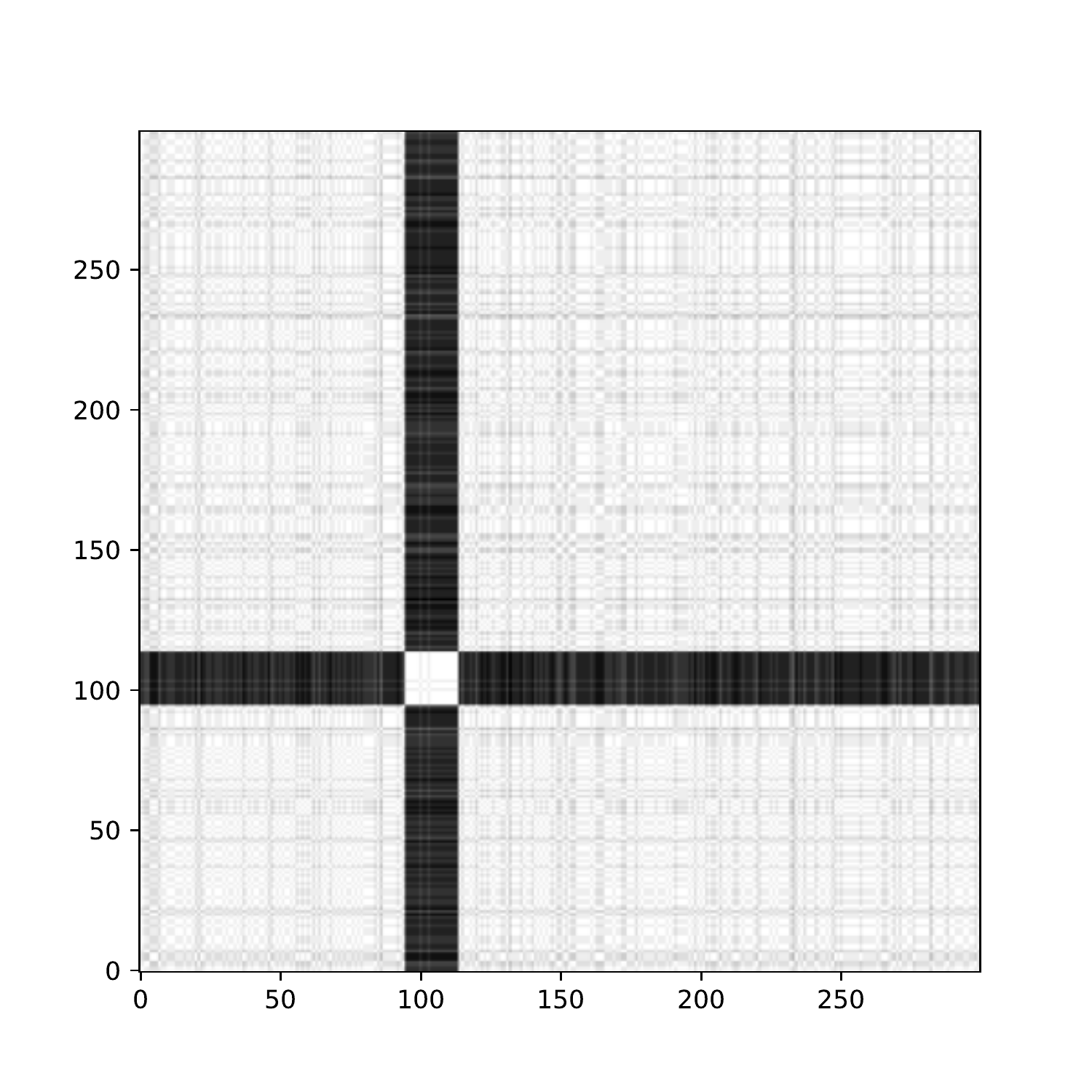}
	}%
	\subfloat[Gramian angular fields transformation\label{fig:example:step-recovery:gaf}]{%
		\includegraphics[width=.5\linewidth]{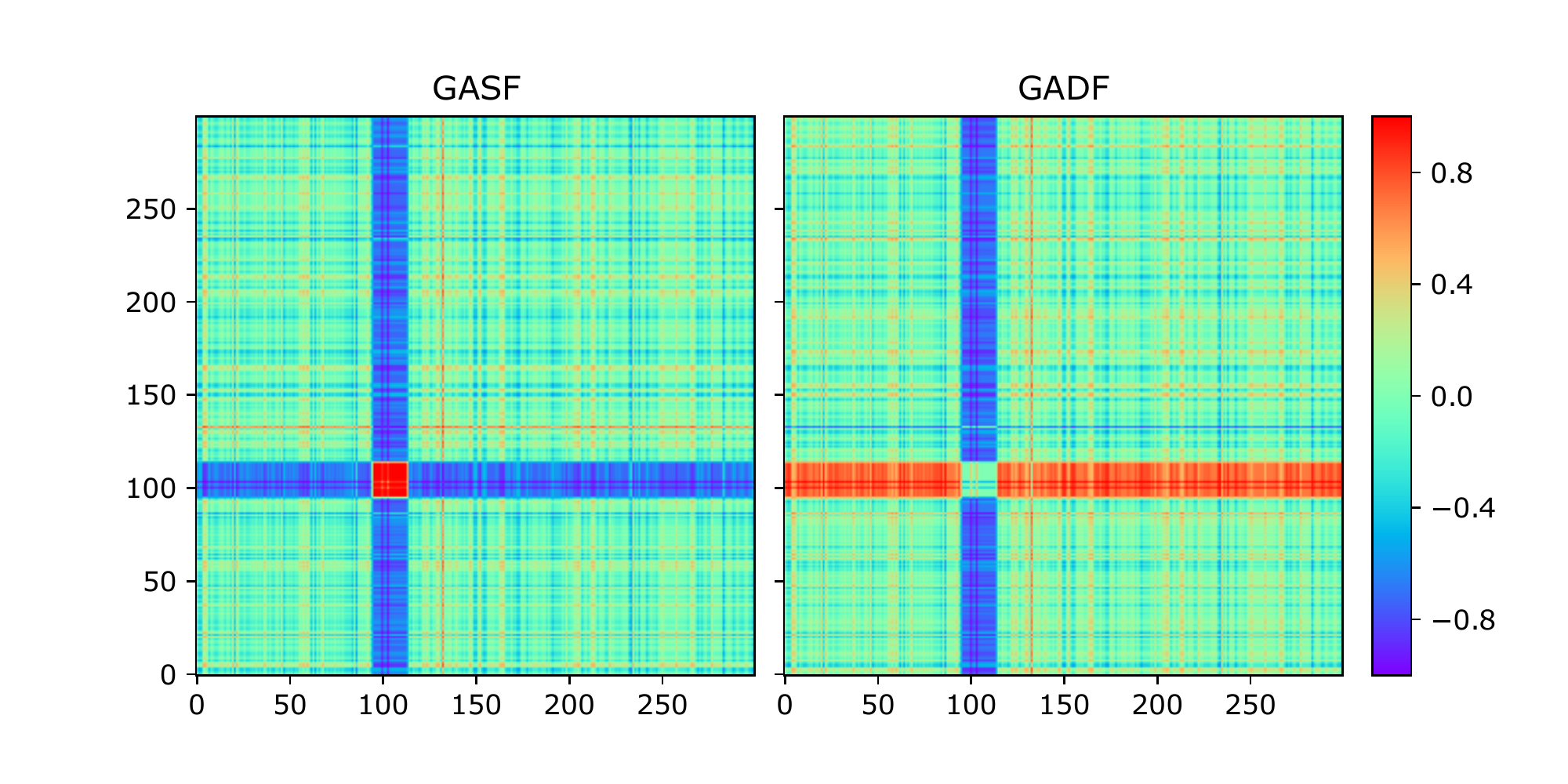}
	}%
	\caption{Distinct representations of the link layer RSSI measurement data for sudden degradation with recovery anomaly (SuddenR).}
	\label{fig:example:step-recovery}
\end{figure*}

\begin{figure*}[htbp]
	\centering
	\subfloat[Time-value perspective\label{fig:example:spikes:ts}]{
		\includegraphics[width=.25\linewidth]{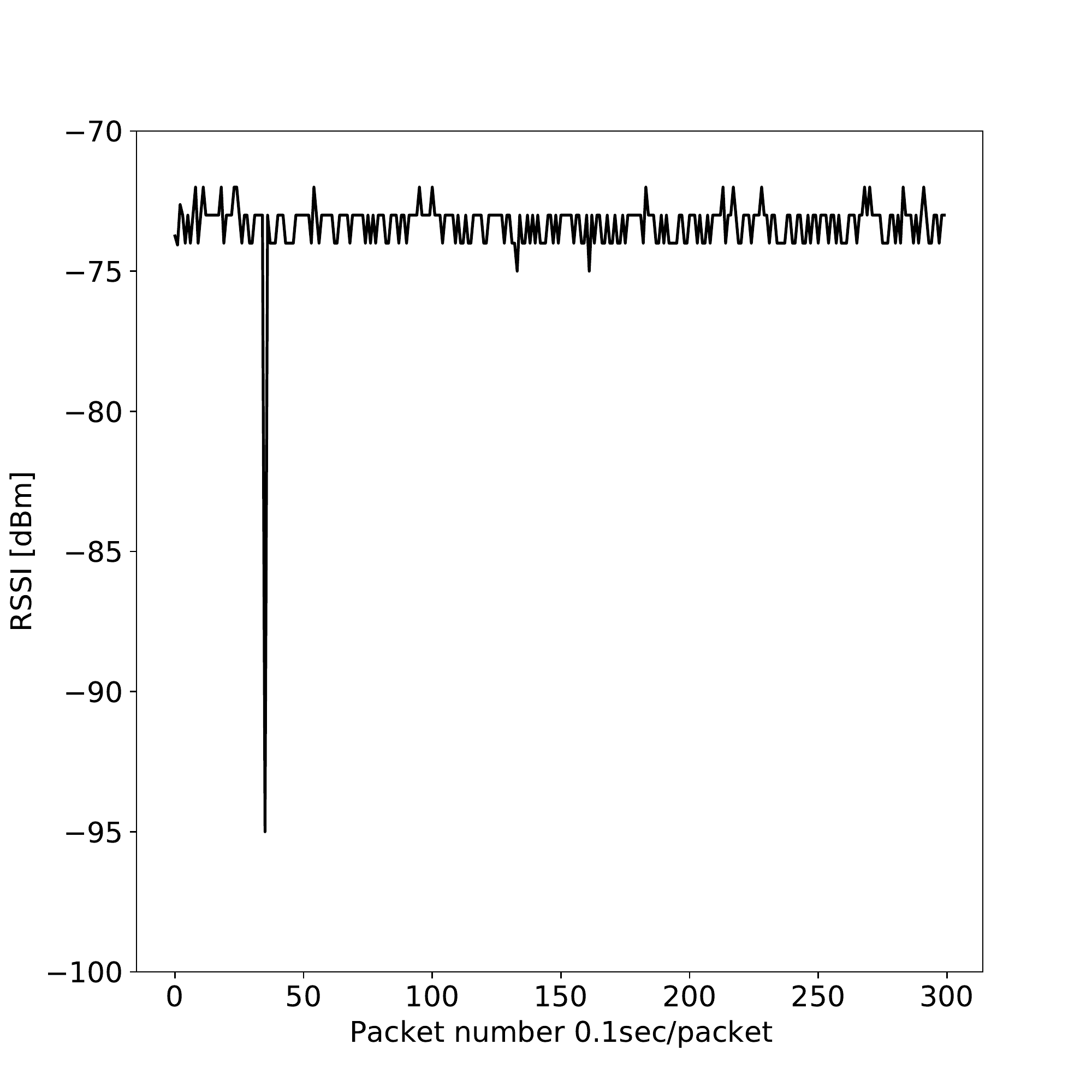}
	}%
	\subfloat[Recurrence plot transformation\label{fig:example:spikes:rp}]{
		\includegraphics[width=.25\linewidth]{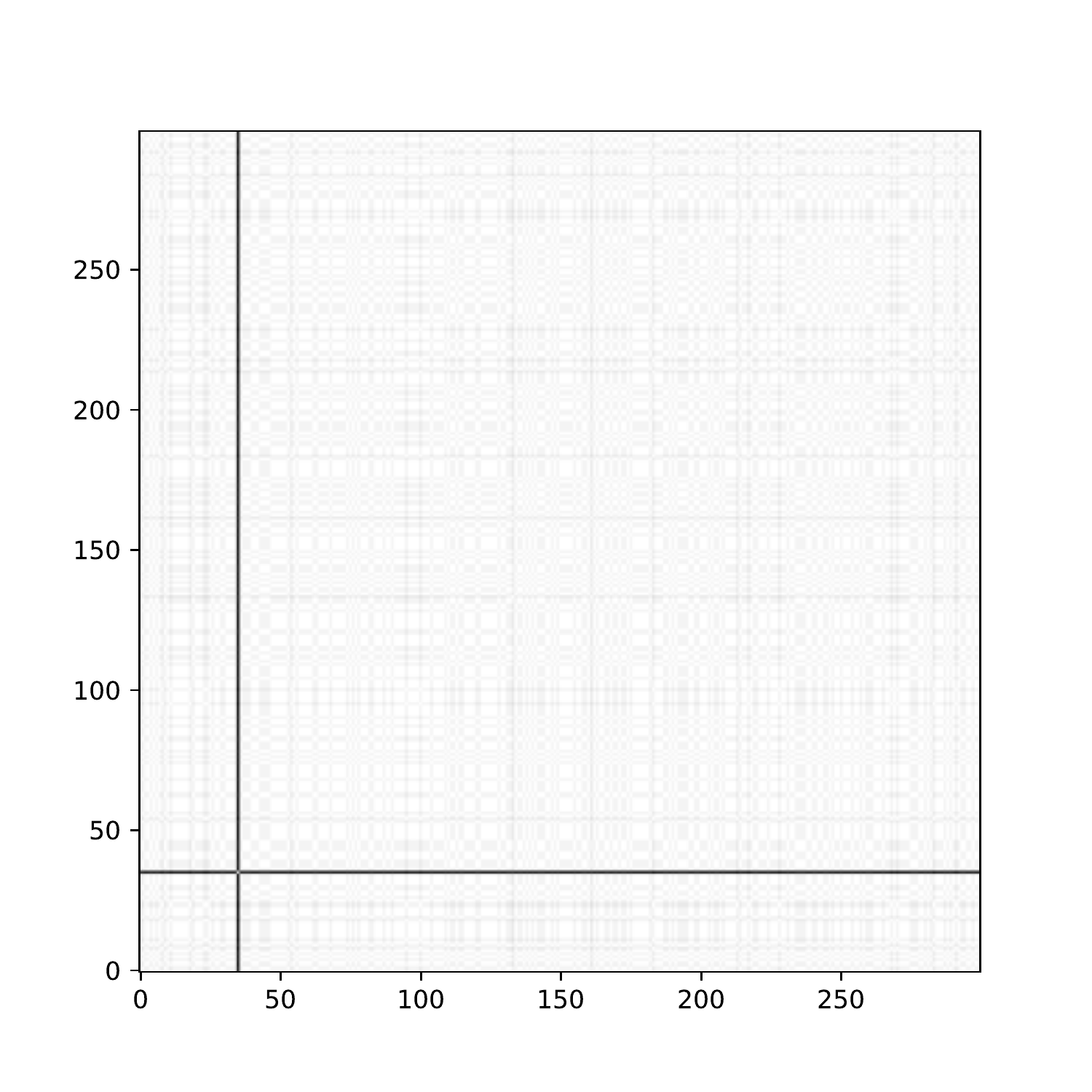}
	}%
	\subfloat[Gramian angular fields transformation\label{fig:example:spikes:gaf}]{%
		\includegraphics[width=.5\linewidth]{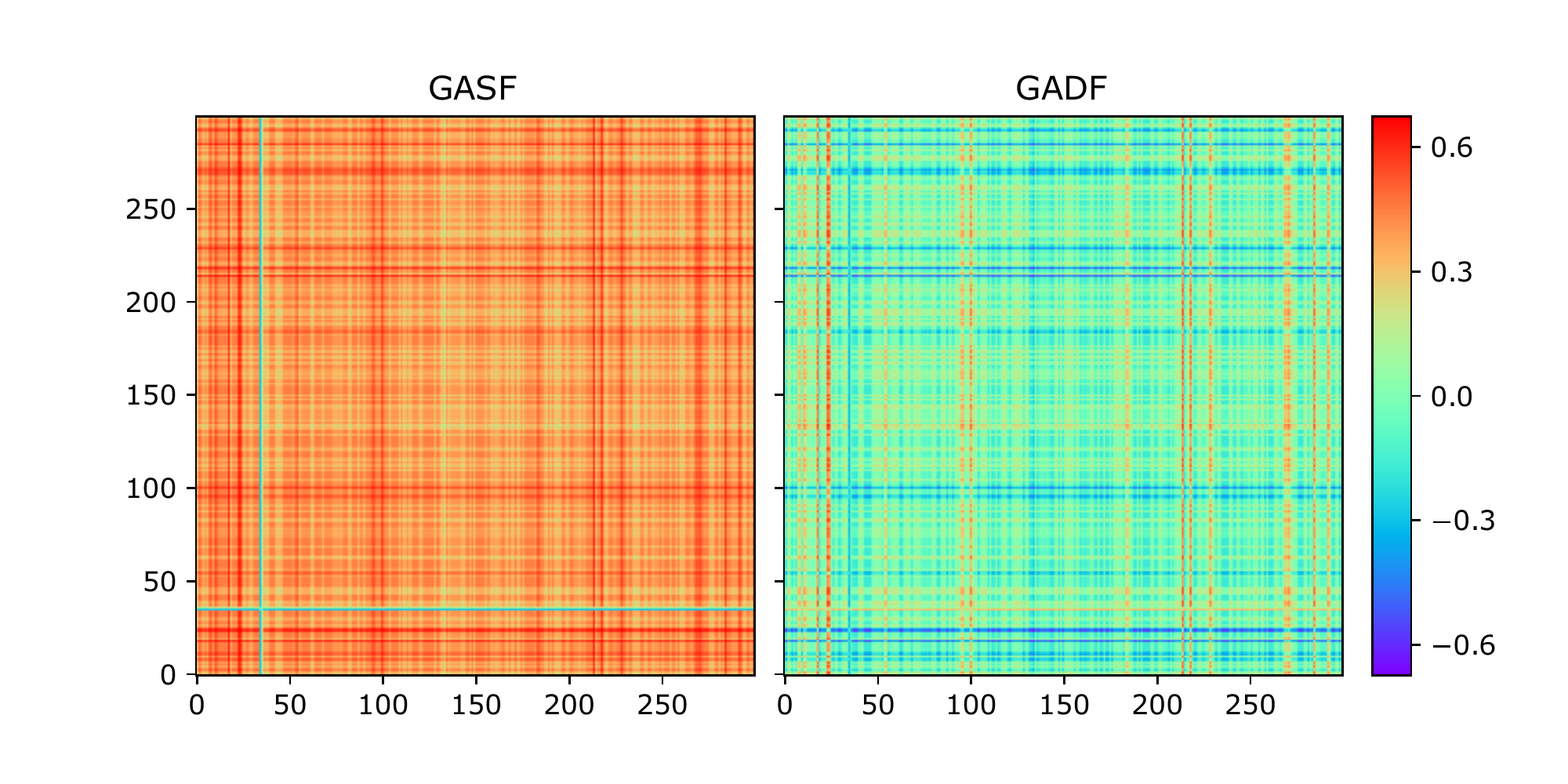}
	}%
	\caption{Distinct representations of the link layer RSSI measurement data for spike-like instantaneous degradation anomaly (InstaD).}
	\label{fig:example:spikes}
\end{figure*}

\begin{figure*}[htbp]
	\centering
	\subfloat[Time-value perspective\label{fig:example:slow:ts}]{
		\includegraphics[width=.25\linewidth]{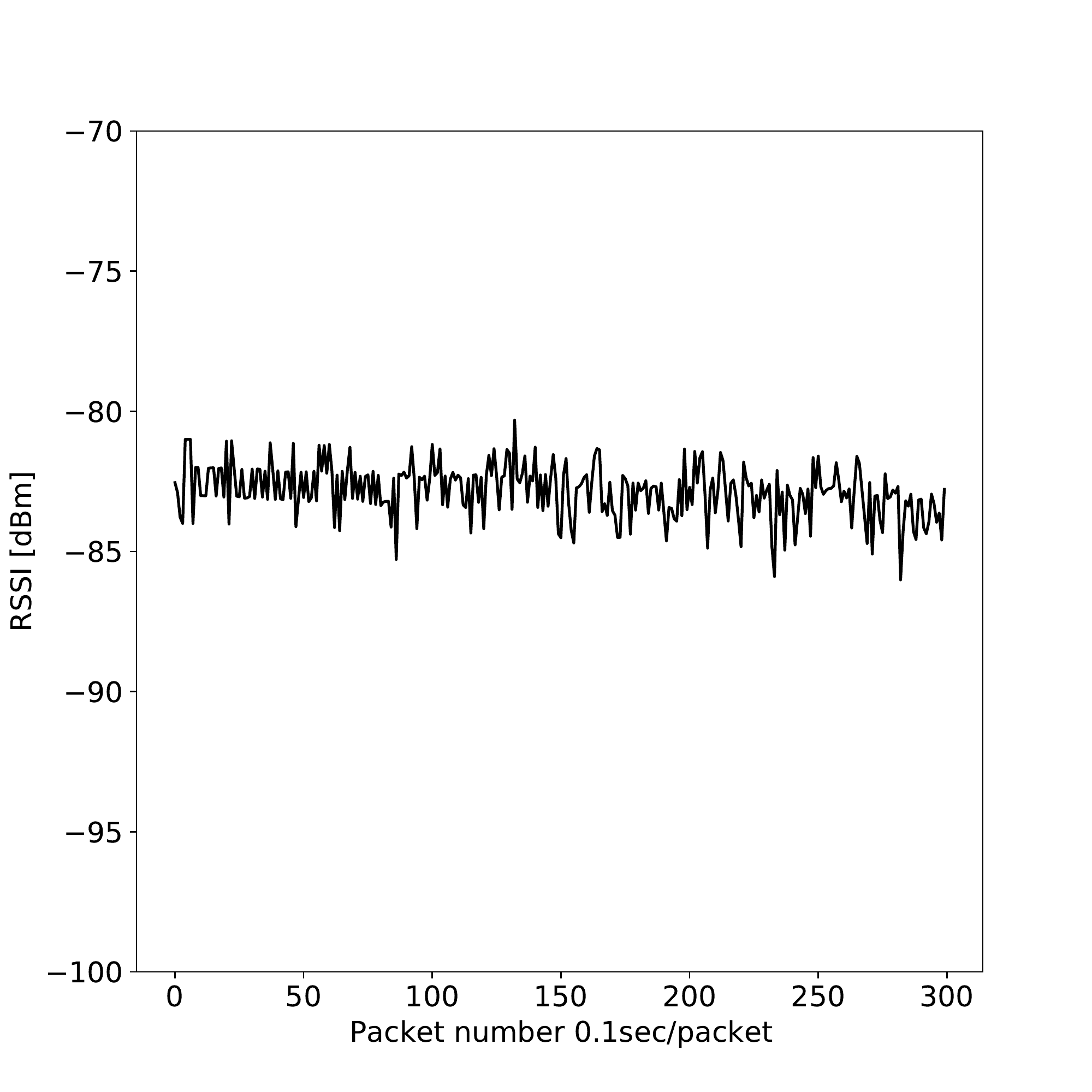}
	}%
	\subfloat[Recurrence plot transformation\label{fig:example:slow:rp}]{
		\includegraphics[width=.25\linewidth]{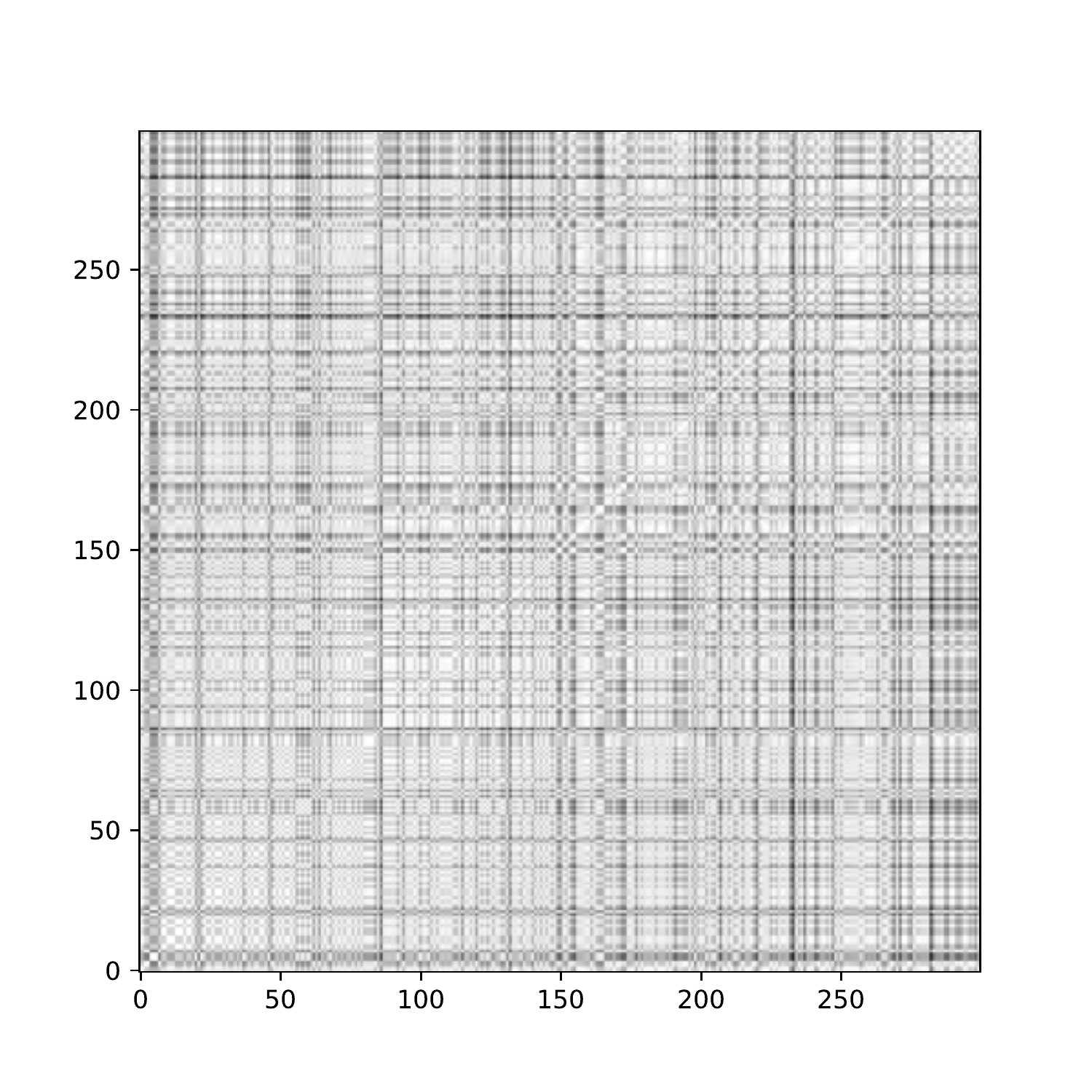}
	}%
	\subfloat[Gramian angular fields transformation\label{fig:example:slow:gaf}]{
		\includegraphics[width=.5\linewidth]{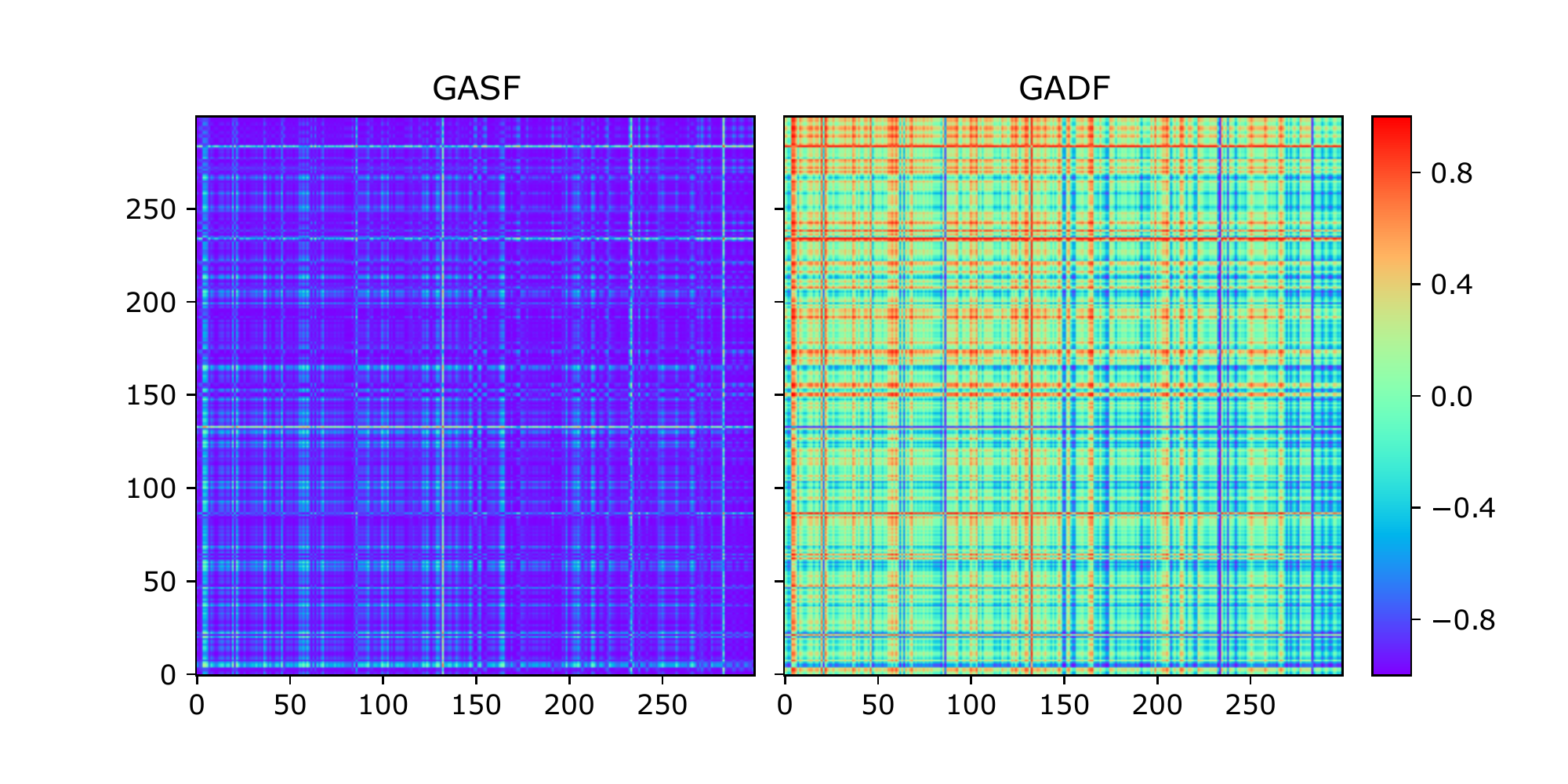}
	}%
	\caption{Distinct representations of the link layer RSSI measurement data for slow degradation anomaly (SlowD).}
	\label{fig:example:slow}
\end{figure*}

For the case of the five class problem, the first anomaly, called Sudden degradation without recovery (SuddenD), is an anomaly where the signal unexpectedly drops to a minimal value and never recovers as depicted in Figure~\ref{fig:example:norecovery:ts}. The second anomaly, called Sudden degradation with recovery (SuddenR), has a certain similarity to SuddenD only that the signal after a certain period of time recovers from minimal back to the normal value as in Figure~\ref{fig:example:step-recovery:ts}. The third anomaly that was defined was Instantaneous degradation (InstaD) that shows itself as a spike within a trace. Here the value of a trace drops to a minimal for a very short, instantaneous, amount of time and then recovers back to a normal value as in Figure~\ref{fig:example:spikes:ts}. The fourth anomaly that was defined is a Slow degradation anomaly (SlowD). This anomaly shows itself in slightly decreasing slope within a trace, where values slowly but gradually decrease like it can be seen in Figure~\ref{fig:example:slow:ts}.
\subsection{Time-series transformation}
\begin{figure*}[htbp]
	\centering
	\subfloat[Time-value perspective\label{fig:example:noanomaly:ts}]{
		\includegraphics[width=.25\linewidth]{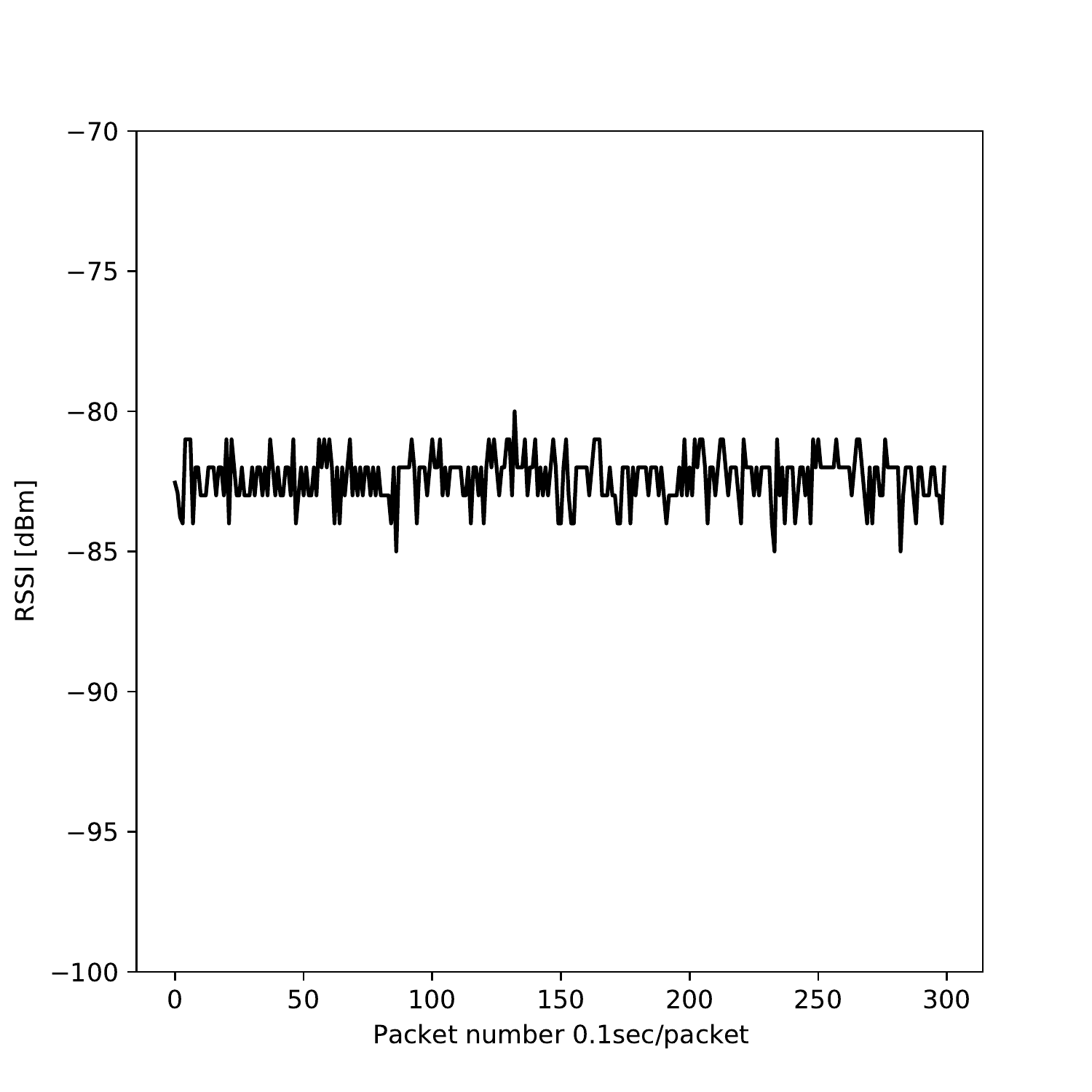}
	}%
	\subfloat[Recurrence plot transformation\label{fig:example:noanomaly:rp}]{
		\includegraphics[width=.25\linewidth]{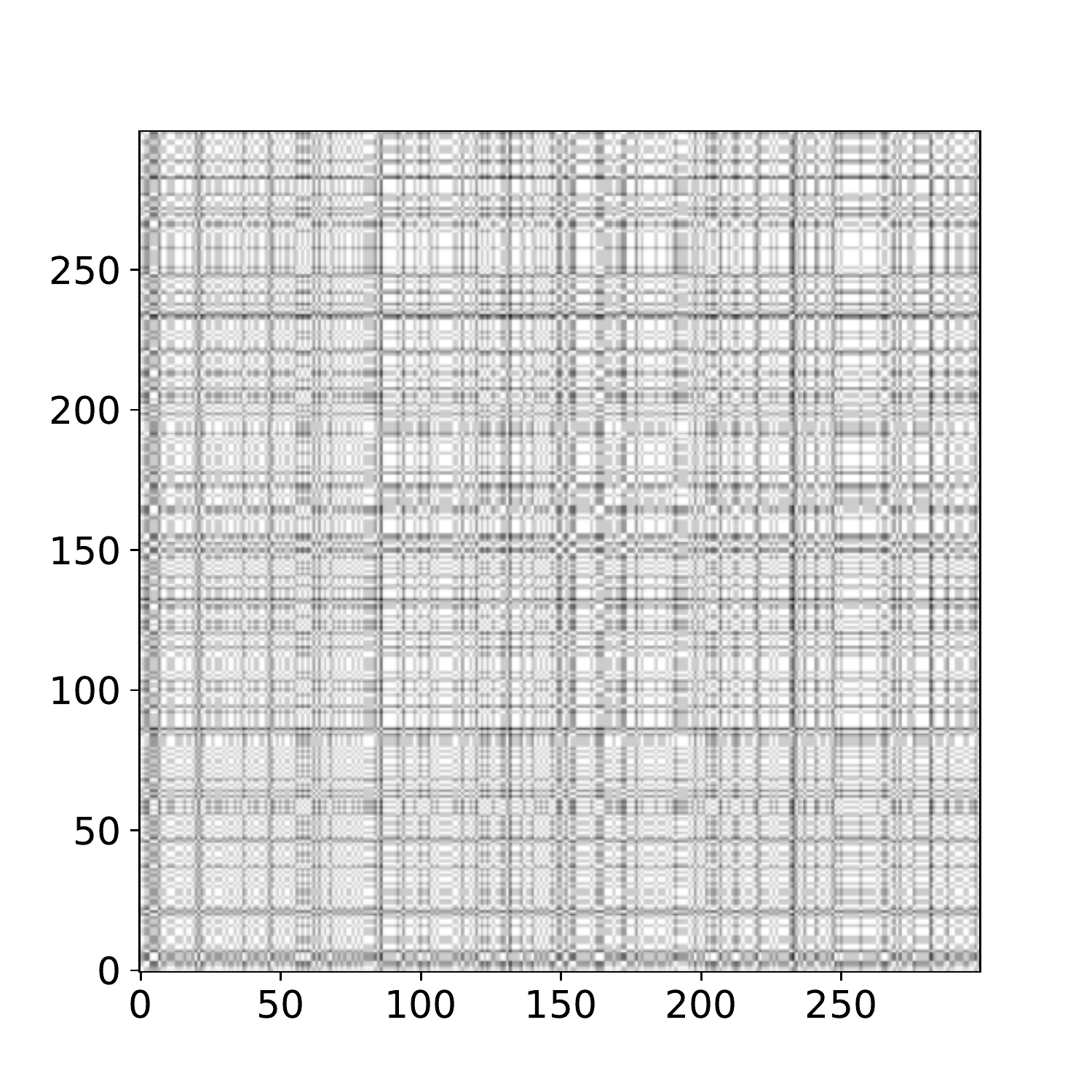}
	}%
	\subfloat[Gramian angular fields transformation\label{fig:example:noanomaly:gaf}]{%
		\includegraphics[width=.5\linewidth]{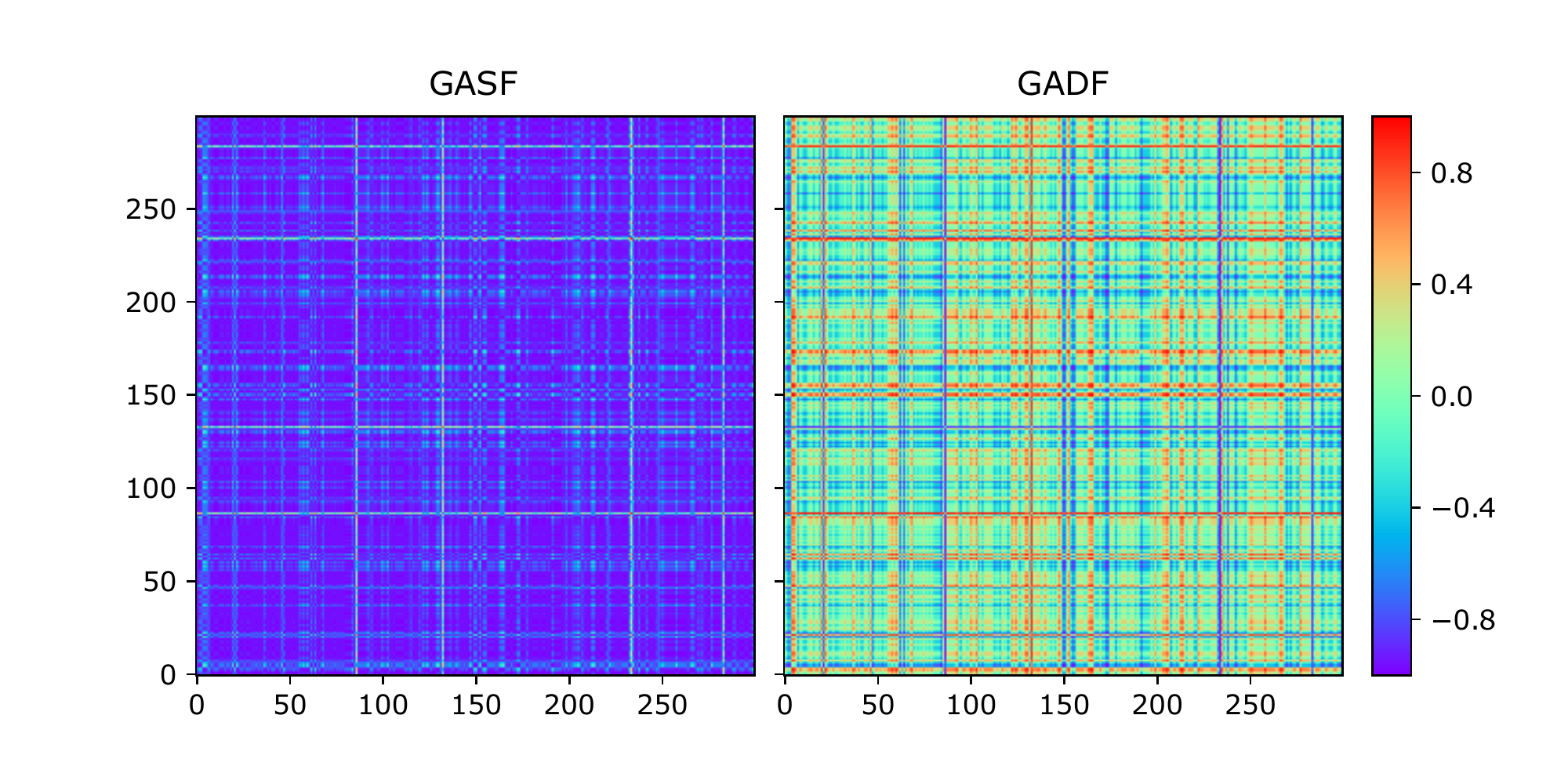}
	}%
	\caption{Distinct representations of the link layer RSSI measurement data without anomaly.}
	\label{fig:example:noanomaly}
\end{figure*}
We define the transformation T depicted in Figure \ref{fig:problem_formulation}  as a function that transforms the input time series S to the tensor H from Eq. \ref{mathref:classification} as:
\begin{equation}
	\label{mathref:transformation}
	H=T(S)
\end{equation}
For multidimensional time series (TS) of dimension K, S takes the form of a $[S]_{K\times N}$ matrix and represents a set of K time series traces with length $N$. In this paper we consider a unidimensional time series where $K=1$, thus $[S]_{1\times N}$. The transformation function $T$ can be represented by the identity function, in which case the result $H$ is the same as input time series trace or can represent more complex transformation functions as the ones proposed in this work that transform the time series into images.
\section{Time-series to image transformation}
\label{sec:TS_transform}
Section~\ref{sec:problem_formulation} provided general definition of time series transformation. In the following, we provide two distinct ways to transform time series data representation into images that can be used as features for training deep learning models.
\subsection{Recurrence plot transformation}
\label{sec:RP}
The recurrence plot (RP) is a technique for nonlinear data analysis, that represents a visualisation of a square matrix in which elements correspond to those times steps when parts of a time series are the most similar to each other. The RP transformation takes as its input the matrix representation of unidimensional time series data $S_{1\times N}$, that can be also represented as a vector $\vec{s_N}$. This can be seen in the Equation~\ref{mathref:timeseries}, where $N$ is the length of the time series. According to Equation~\ref{mathref:rplot:distance}, the absolute value of the difference between two points is computed and represents the distance between the points, which is then subtracted from the predefined threshold $\epsilon$. The result $h$ is then converted to binary values through the use of the Heaviside function $\Theta$ presented in the Equation~\ref{mathref:rplot:heaviside}. The final result is an image $H$ of size $N\times N$. In special cases the threshold and Heaviside function may be omitted resulting in a non binarized matrix of distances between points in the time series.
\begin{equation}\label{mathref:timeseries}
S_{1\times N} =\vec{s_N}= (s_1, s_2,..., s_N)
\end{equation}

\begin{equation}
	\label{mathref:rplot:distance}
	h = \epsilon - 
	\begin{pmatrix}
		\Vert s_N - s_1 \Vert & \Vert s_N - s_2 \Vert & ... &  \Vert s_N - s_N \Vert\\
		\vdots & \vdots & \ddots & \vdots\\
		\Vert s_2 - s_1 \Vert & \Vert s_2 - s_2 \Vert & ... &  \Vert s_2 - s_N \Vert\\
		\Vert s_1 - s_1 \Vert & \Vert s_1 - s_2 \Vert & ... &  \Vert s_1 - s_N \Vert\\
	\end{pmatrix}
\end{equation}

\begin{equation}\label{mathref:rplot:heaviside}
	H = \Theta(h)
\end{equation}
A trace without an anomaly can be seen in Figure~\ref{fig:example:noanomaly:rp}. No pattern is apparent, as the arrangement within the image can be labelled as random and correlates with the randomness of the time series representation of the trace. On the other hand, Figure~\ref{fig:example:norecovery:rp} depicts a recurrence plot of the SuddenD anomaly. The typical representation of this anomaly is a white rectangle in the upper right corner. The lower left corner of this rectangle represents the time sample where the anomaly occurred, while the width of the rectangle is the same as the width of the anomaly of the time series anomaly represented in Figure~\ref{fig:example:norecovery:ts}.

Looking at Figure~\ref{fig:example:step-recovery:rp}, the SuddenR anomaly appears as a cross that has a small white rectangle in its centre somewhere along the diagonal from the lower left corner to the upper right corner of the image. The lower left corner of the white square represents the beginning of the anomaly, while the upper right corner of the white square shows where the recovery occurred. The size of the rectangle depends on the width of the anomaly.

The third type of anomaly is depicted in Figure~\ref{fig:example:spikes:rp}. The InstaD anomaly is very similar to the SuddenR anomaly, except that it is much narrower, which is due to its short occurrence within a trace. Just like SuddenR, this anomaly can be observed along the diagonal line from the lower left corner to the upper right corner.

Finally, in Figure~\ref{fig:example:slow:rp}, the SlowD anomaly can be observed. This anomaly does not have a typical representation. In some images it is seen as an area of higher point density along the diagonal from the bottom left to the top right, some have higher density at the top and right edge of the image, while some other images may have a similar distribution as traces without an anomaly. As such, this anomaly with RP transform is harder to detect with the naked eye than others.
\subsection{Gramian angular field transformation}
\label{sec:GAF}
The Gramian angular field is a transformation of a time series that represents the temporal correlation between points within a time series. The end result is a square image representation of the input time series. This approach consists of two techniques, one is the Gramian angular summation field (GASF) and the other is the Gramian angular difference field (GADF). Both techniques are computed in a similar way. First, the time series needs to be scaled with a min-max normalization and then transformed to a polar coordinate system. The angles $\phi_N$ from the polar plot are then used to compute GASF and GADF. For the sum field, the angular cosine function of the sum between two points is computed, which is represented in Equation \ref{mathref:GASF}, where $H_S$ represents the GASF transformation. For the GADF representation, the angular sine of the difference between each two points is computed as shown in Equation \ref{mathref:GADF}, where $H_D$ represents the GADF transform. Both GAF representations look similar to those transformed with recurrence plot. The size of the transformed image is $N\times N$, where $N$ represents the length of the time series used in the transformation. 
\begin{equation}
	\label{mathref:GASF}
	H_S =
	\begin{pmatrix}
		\cos(\phi_N + \phi_1) & \cos(\phi_N + \phi_2) & ... & \cos(\phi_N + \phi_N)\\
						  \vdots & \vdots & \ddots & \vdots\\
		\cos(\phi_2 + \phi_1) & \cos(\phi_2 + \phi_2) & ... & \cos(\phi_2 + \phi_N)\\
		\cos(\phi_1 + \phi_1) & \cos(\phi_1 + \phi_2) & ... & \cos(\phi_1 + \phi_N)\\
	\end{pmatrix}
\end{equation}

\begin{equation}
	\label{mathref:GADF}
	H_D =
	\begin{pmatrix}
		\sin(\phi_N - \phi_1) & \sin(\phi_N - \phi_2) & ... & \sin(\phi_N - \phi_N)\\
		\vdots & \vdots & \ddots & \vdots\\
		\sin(\phi_2 - \phi_1) & \sin(\phi_2 - \phi_2) & ... & \sin(\phi_2 - \phi_N)\\
		\sin(\phi_1 - \phi_1) & \sin(\phi_1 - \phi_2) & ... & \sin(\phi_1 - \phi_N)\\
	\end{pmatrix}
\end{equation}
A trace without an anomaly can be seen in the Figure~\ref{fig:example:noanomaly:gaf}. Similar to the  recurrence plot transformation, there is no obvious pattern in the images and the arrangement also looks random. On the other hand, it can be seen that in Figure~\ref{fig:example:norecovery:gaf} that depicts the transformation corresponding to the  SuddenD anomaly, a rectangle in the upper right corner of the generated image indicates the anomaly. It can be noticed that the representation is similar to the one generated by the recurrence plot and that it is more visible in the GASF plot rather than the GADF.

Figure~\ref{fig:example:step-recovery:gaf} depicts the GAF transforms for the SuddenR anomaly. The anomaly appears as a small rectangle somewhere along the diagonal of the image. The width of the rectangle is the same as the width of the anomaly within the time series as can be seen by comparing with the width of the time series anomaly represented in Figure~\ref{fig:example:step-recovery:ts}. 

InstaD can be seen in Figure~\ref{fig:example:spikes:gaf}. It manifests as a small green cross along the diagonal within the GASF image, while it is harder to spot in GADF. Similar to all other GAF anomaly transformations, this one also bares a resemblance to RP representation, only that it is less distinct.

The final GAF anomaly representation is SlowD, which is shown in Figure~\ref{fig:example:slow:gaf}. This is best observed in GADF, where it can be seen that the highest values in the upper left corner of the image. This represents that the values at the beginning of the curve trace are much more similar to each other than the values at the end of the trace. No obvious pattern can be observed in the GASF image.
\begin{figure*}[thb]
	\centering
	\includegraphics[width=0.8\linewidth]{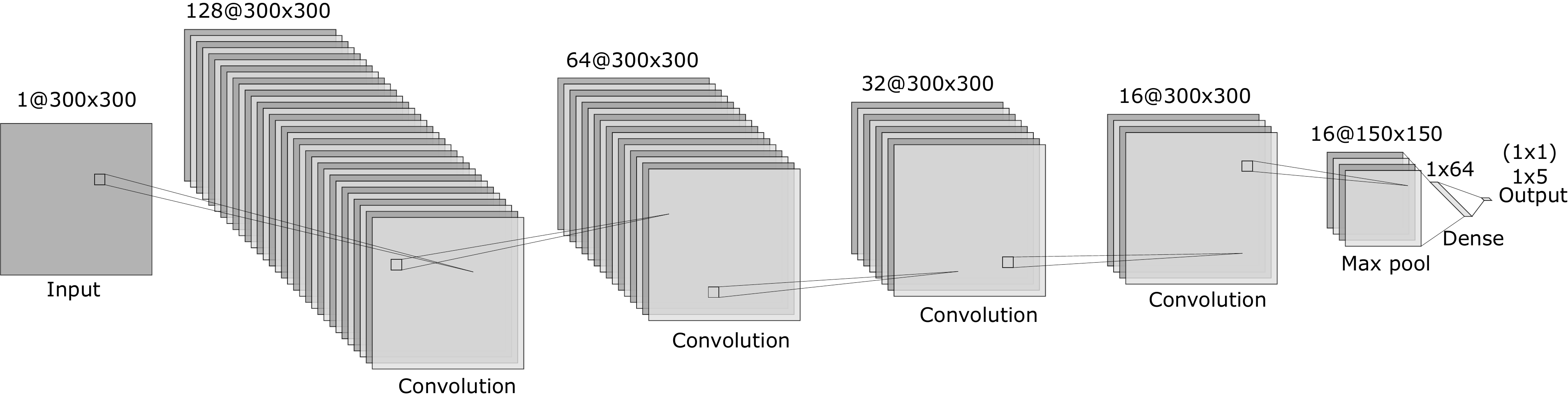}
	\caption{Deep learning network model for multi-class and binary detection for GAF and RP}
	\label{fig:model:RP_GAF_multi}
\end{figure*}
\section{Proposed DNN classification model}
\label{sec:DL_models}
We define our classification model as function $\Phi$ depicted in Figure~\ref{fig:problem_formulation} that transforms the input transformed data $H$ to the set of target classes $C$ as provided in Equation~\ref{mathref:dnn1}. 
\begin{equation}
	\label{mathref:dnn1}
C=\Phi (H) = \Psi (Z), \textrm{where }Z = W\cdot H + B	
\end{equation}
where $Z$ takes the input data $H$ and multiplies it with weights and adds bias from the model to it. The activation function $\Psi$ is then applied to $Z$ and the predicted class $C$ is returned.

We have designed two architectures of convolutional neural networks $\Phi (H)$, one for binary and the other for the five-class classification. In essence, the same architecture is used for both classification problems, the only difference being the output layer, which is adapted according to the classification type.  The network architecture was designed using an iterative process using the following design considerations:

\subsection{Design considerations}
\label{subsec:considerations}
 In order to ensure that our architecture is mindful of resources, we considered studies on estimating the energy consumption of ML models~\cite{Garcia2019} that show that the increasing complexity of models, manifested in the number of weights, the type of layers and their respective parameters, affects both their performance and energy efficiency. We followed the following design steps:

 \paragraph{Reduce the number of the layers of the network} The computational complexity of a network is typically evaluated as the number of floating point operations (FLOPs) needed to make a prediction. This depends on the number of layers $L$ and the computational complexity of each individual layer $F_\text{l}$ as per Eq.\ref{eqn:ModelFlops}. During our iterative design process, we considered $L\in\{5, ..,20\}$.
	\begin{equation}
	\label{eqn:ModelFlops}
	M_{FLOPs} = \sum_{l=1}^{L} F_\text{l} ,
	\end{equation}
	
 \paragraph{Optimize the convolutional layers.} Convolutional layers represent a sequence of matrices used to extract features from the image. A convolutional layer consists of a set of filters of size $K_\text{r} \times K_\text{c}$ used to scan an input tensor of size $I_\text{r} \times I_\text{c} \times C$ with a stride $S$. More precisely, the number of all FLOPs per filter $F_\text{pf}$ is given by Eq.~\ref{eqn:FlopsPerFilter}.
\begin{equation}
\label{eqn:FlopsPerFilter}
F_\text{pf} = (\frac{I_\text{r} - K_\text{r} + 2P_\text{r}}{S_\text{r}} + 1)(\frac{I_\text{c} - K_\text{c} + 2P_\text{c}}{S_\text{c}} + 1) (2CK_\text{r}K_\text{c} + 1)
\end{equation}

 The first term of the equation gives the height of the output tensor, where $I_\text{r}$ is the size of the input rows, $K_\text{r}$ is the height of the filter, $P_\text{r}$ is the padding and $S_\text{r}$ is the size of the stride. The second term represents the same calculation for the width of the output tensor, where the indices in  $I_\text{c}$, $K_\text{c}$, $P_\text{c}$ and $S_\text{c}$ correspond to the input columns. The last term provides the number of computations per filter for each of the input channels $C$ that represent the depth of the input tensor and the bias. 

 The number of FLOPs used throughout the convolutional layer is equal to the number of filters times the flops per filter given in Eq.~\ref{eqn:FlopsPerFilter}, i.e. $F_\text{c}=(F_\text{pf} + N_\text{ipf})N_\text{f}$. However, in the case where ReLU are used, one additional comparison and multiplication are required to calculate the number of FLOPs used in one epoch $F_\text{pe}$. We therefore added the number of FLOPs used for each filter and the number of instances for each filter and then multiplied by the number of all filters $N_\text{f}$:  
\begin{equation}
\label{eqn:FlopsLayerRelu}
F_\text{c} = (F_\text{pf} + (2CK_\text{r}K_\text{c} + 1))N_\text{f} .
\end{equation}

 During our iterative design process, we considered:
\begin{itemize}
	\item  Optimizing the number of filters $N_{f}\in\{16, 32, ...,128\}.$ from Eq. \ref{eqn:FlopsLayerRelu}.
	\item  Optimizing the kernel size $K_{r}=K_{c}\in\{2, 3, ..., 7\}.$ from Eq. \ref{eqn:FlopsPerFilter}.
\end{itemize}

\subsection{Architecture}
\label{subsec:architecture}
 The proposed architecture depicted in Figure~\ref{fig:model:RP_GAF_multi}  yielded the best performance/resource utilization during the design process. On the left side of the figure there is an input of size $300\times300$ pixels, which is the size of the image resulting from transforming the TS. The image is then fed into the neural network, which starts with four convolutional layers. The first layer uses 128 filters with kernel size $3\times3$ pixels. The next three convolutional layers use 64, 32, and 16 filters with a kernel size of $7\times7$. After the last convolution layer, max-pooling is applied to the output, reducing the height and width of the output by half.  The data is then flattened and fed into the dense layer, consisting of 64 nodes, and connected to the output layer, which is either the size of 1 or 5 depending on the classification type. All hidden layers use the ReLU function that inserts non-linearity into the model and thus  helping with with the classification of classes with non-linear boundary. Finally, the output layer uses a sigmoid activation function that is suitable also for multi-class classification.

\section{Methodology and experimental details}
\label{sec:methodology}
 To develop and evaluate our model we follow the methodology presented in the following subsections. Experimental settings such as the train/test split and the energy consumption od the model are provided within the respective subsections.
\subsection{Dataset generation}
For our experimental evaluation, we choose the Rutgers dataset~\cite{rutgers-noise} containing real-world measurements and then we synthetically injected anomalies. These traces were then converted into RP, GASF and GADF images resulting in 3 datasets for solving our binary and five-class classification problem.
\begin{table}[htbp]
	\centering
	\ra{1.2}
	\begin{threeparttable}[b]
		\caption{Synthetic anomaly injection method.}
		\label{tab:injection-scenario}
		\begin{tabularx}{\linewidth}{@{}lllll@{}}
			\toprule
			Type
			& Links
			& Affected
			& Appearance
			& Persistence
			\\\midrule
			
			SuddenD 
			& \multirow{4}{*}{2\,123}
			& \multirow{4}{*}{33\% (700)}
			& once, [200$^\textrm{th}$,~280$^\textrm{th}$] 
			& for $\infty$
			\\
			
			SuddenR 
			&
			&
			& once, [25$^\textrm{th}$,~275$^\textrm{th}$]
			& for [5,~20]
			\\
			
			InstaD 
			&
			&
			& on $\approx$1\% of a link
			& for 1 datapoint
			\\
			
			SlowD 
			&
			&
			& once, [1$^\textrm{st}$,~20$^\textrm{th}$]
			& for [150, 180]$^\dagger$
			\\
			
			\bottomrule
		\end{tabularx}
		\begin{tablenotes}
			\item[$\dagger$] RSSI$(x, \textrm{start})$ $\leftarrow$ RSSI$(x)$ + $\min(0, -\textrm{rand}(0.5, 1.5)\cdot(x-\textrm{start}))$
		\end{tablenotes}	
	\end{threeparttable}
\end{table}

The Rutgers dataset consists of link traces from 29 nodes at 5 different noise levels. The dataset contains the raw Received Signal Strength Indicator (RSSI), sequence numbers, source node ID, destination node ID and artificial noise levels. Each RSSI value represents the signal strength of a received packet sent every 100 miliseconds, for a period of 30 seconds. Thus, we obtain traces with a length of 300 RSSI samples. 

To inject the anomalies into the Rutgers dataset, we first filtered out samples that did not exhibit packet loss. This left 2123 samples, of which 33\%,  same as in~\cite{cerar2020anomaly}, were randomly injected with some kind of anomaly while the other links were left untouched. The anomalies were randomly injected according to~\cite{cerar2020anomaly} according to the parameters from Table~\ref{tab:injection-scenario}. This gave us 4 raw time series datasets, one for each of the four anomalies, which together constitute our final dataset. It consists of 8492 samples, with each anomaly represented by 700 samples, while there are a total of 5692 traces that have no anomaly. These traces are then transformed into RP, GASF, GADF images and  snapshots of raw time-series resulting in training instances of size $300 \times 300$. In the RP transformation, the Heaviside function and thresholding were omitted as such design decisions gave the best final results. 

\begin{table*}[!htbp]
	\centering
	\ra{1.2}
	\footnotesize
	\begin{threeparttable}[b]
		\caption{Classification results of various based models (baseline, RP, GASF and GADF).}
		\label{tab:results-classification}
		\begin{tabular}{lllllllllllllllll}
			\toprule
			
			\multirow{2}{*}{Classification}
			&\multirow{2}{*}{Class}
			& \multicolumn{3}{c}{Baseline TS snapshot}
			& \phantom{}
			& \multicolumn{3}{c}{recurrence plot}
			& \phantom{}
			& \multicolumn{3}{c}{GASF}
			& \phantom{}
			& \multicolumn{3}{c}{GADF}
			\\\cmidrule{3-5}\cmidrule{7-9}\cmidrule{11-13} \cmidrule{15-17}
			
			& 
			& Prec.
			& Rec.
			& F1
			& 
			& Prec.
			& Rec.
			& F1
			& 
			& Prec.
			& Rec.
			& F1
			& 
			& Prec.
			& Rec.
			& F1
			\\\midrule

			\multirow{2}{*}{Binary} & True	& 0.90 & 0.90 & 0.90 &	& 0.99 & 1.00 &  \textbf{0.99} &	& 0.97 & 0.98 & 0.98 &	& 0.91 & 0.92 & 0.91\\
			
			& False	& 0.60 & 0.72 & 0.65 &	& 0.99 & 0.96 &  \textbf{0.97} &	& 0.94 & 0.93 & 0.94 &	& 0.97 & 0.97 &  \textbf{0.97}  \\\midrule
			
			\multirow{5}{*}{Multiclass} & SuddenD	& 0.99 & 0.99 & 0.99 &	& 1.00 & 1.00 &  \textbf{1.00} &	& 0.99 & 0.99 & 0.99 &	& 1.00 & 1.00 &  \textbf{1.00} \\
			
			& SuddenR	& 0.93 & 0.78 & 0.85 &	& 1.00 & 1.00 &  \textbf{1.00} &	& 0.99 & 0.92 & 0.95 &	& 0.98 & 0.99 & 0.98  \\
			
			& InstaD	& 0.66 & 0.70 & 0.68 &	& 0.93 & 0.90 &  \textbf{0.92} &	& 0.92 & 0.88 & 0.90&	& 0.90 & 0.82 & 0.86  \\
			
			& SlowD	& 0.84 & 0.95 & 0.89 &	& 1.00 & 0.99 &  \textbf{0.99} &	& 0.95 & 0.96 & 0.95 &	& 0.75 & 0.99 & 0.85  \\
			
			& No anomaly	& 0.97 & 0.97 & 0.97 &	& 0.99 & 0.99 &  \textbf{0.99}\tnote &	& 0.97 & 0.99 & 0.98 &	& 0.98 & 0.96 & 0.97 \\
			\bottomrule
		\end{tabular}	
	\end{threeparttable}
\end{table*}
\begin{figure*}[!tbh]
	\centering
	\subfloat[SuddenD\label{fig:example:norecovery:XAI}]{
		\includegraphics[width=.23\linewidth]{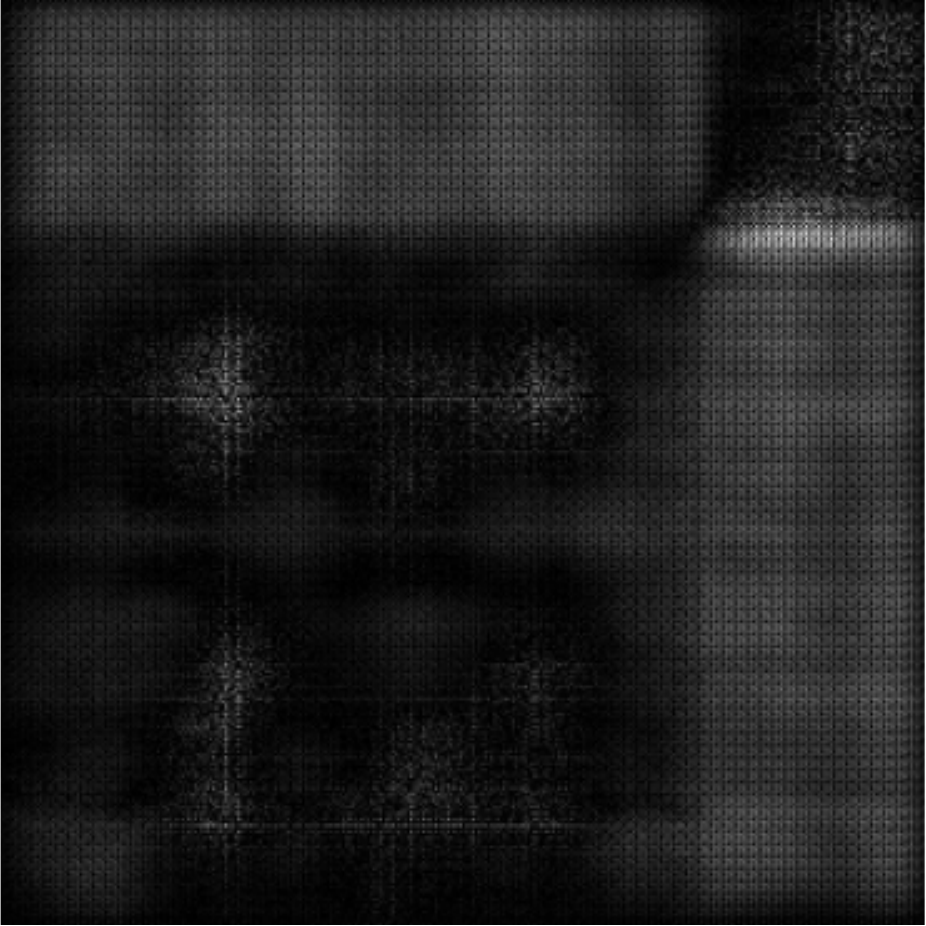}
	}%
	\subfloat[SuddenR\label{fig:example:recovery:XAI}]{
		\includegraphics[width=.23\linewidth]{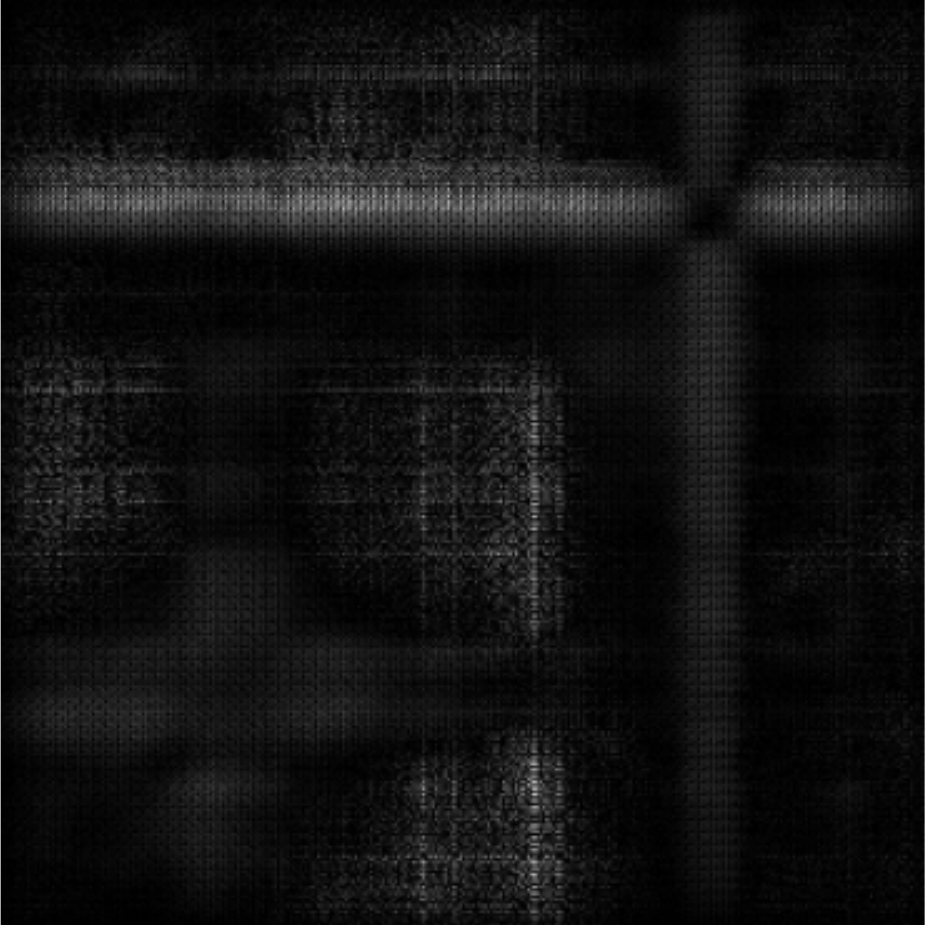}
	}%
	\subfloat[InstaD\label{fig:example:spikes:XAI}]{
		\includegraphics[width=.23\linewidth]{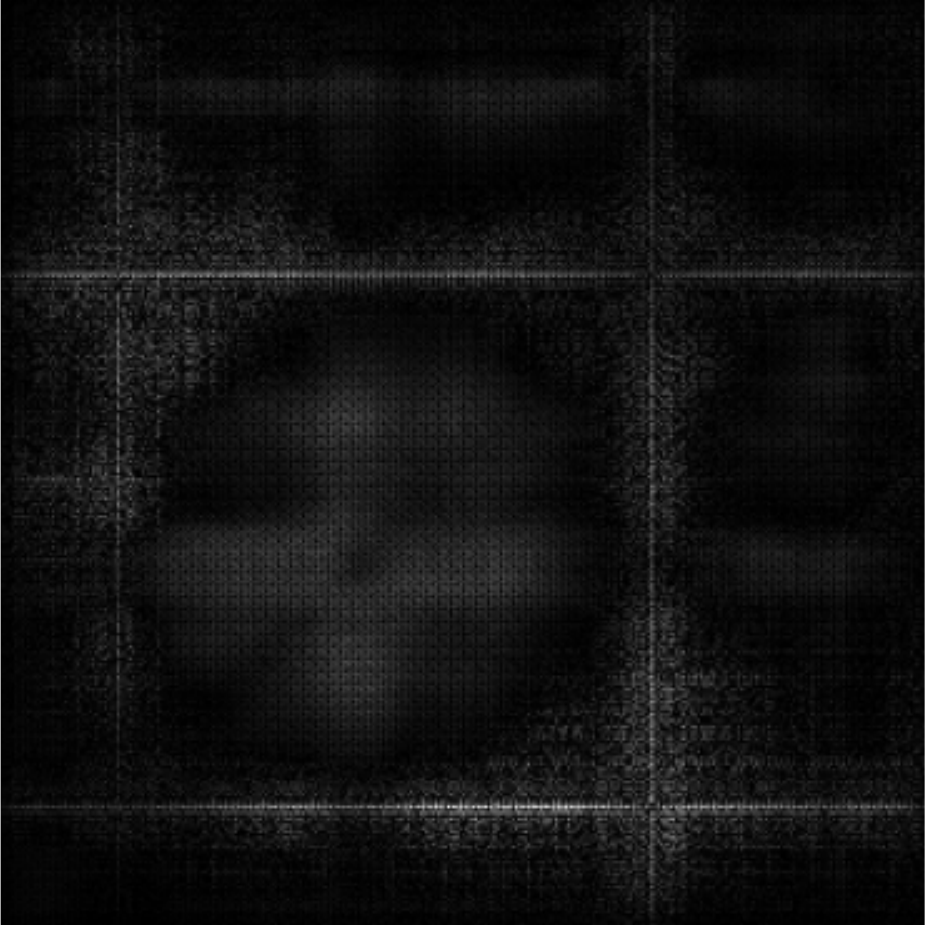}
	}%
	\subfloat[SlowD\label{fig:example:slow:XAI}]{
		\includegraphics[width=.23\linewidth]{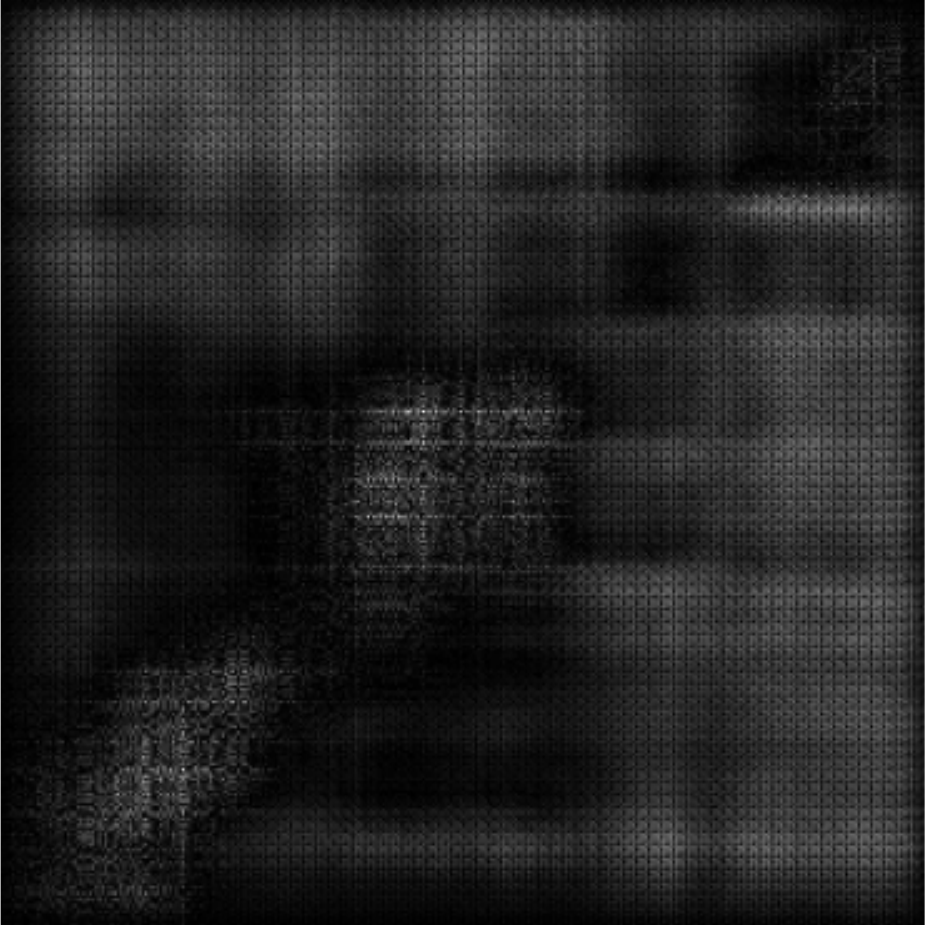}
	}%
	\caption{Distinct representations of explanations of model decisions acquired by Guided Backpropagation.}
	\label{fig:example:XAI}
\end{figure*}

\subsection{Model training}
For training purpose we created a shuffle and split dataset for training and testing with 80:20  split ratio, same as in \cite{cerar2020anomaly} so that the results could be compared. We did the same shuffle split ten times for each model selection. The model is trained on the training set and evaluated on test set so that we can ensure credible results. Because of the imbalanced nature of the dataset, we weighted the classes during training process. The purpose is to penalize the miss-classification made by the minority classes by setting a higher class weight and at the same time reducing weight for the majority class. The "No anomaly" class was our majority class and got a weight of $0.1$, which is lower than the distribution ratio between anomalous and non anomalous links to ensure greater importance to the anomaly samples, while all other classes were assigned a weight of $1$.

\subsection{Model evaluation}
\label{subsec:eval}
To analyze the performance of the considered time-series to image transformations we compared the model designed in Section\ref{sec:DL_models} and trained on Grammian angular fields and recurrence plots to a baseline trained on snapshots of raw time-series data. We also analyze how many positive examples are needed in the dataset to train a good model.

 To compare the proposed model to classical approaches, we consider two techniques widely used for anomaly detection, namely k-nearest neighbours (KNN) and support vector machines (SVM) on raw time series data using the well known and widely used DTW as distance metric.

 As explained in Section \ref{sec:DL_models} we designed a resource-aware classifier for time-series to image datasets. However, there are  well known CNN based DL architectures that are specialized on image classification. While such architectures can be extremely complex, AlexNet~\cite{krizhevsky2012imagenet} and VGG11~\cite{simonyan2014very} are well known and widely used as benchmarks in various communities and application areas. Additionally they also have an architecture that is similar to our proposed model, therefore we selected these two models to compare against.

 Finally, we also compared the performance of the proposed model based on recurrence plots against two state of the art works from the considered application area. More specifically, we compared to the ensemble learner consisting of the best single class classifiers designed by \cite{cerar2020anomaly} and to classifiers designed by \cite{bertalanic_deep_nodate}.

With respect to used metrics, we evaluated the models using the standard per class precision, recall and F1. The precision measures how many instances detected as class A actually belong to class A, expressed as: $\textrm{Precision} = \frac{\textrm{TP}}{\textrm{TP} + \textrm{FP}}$, while recall measures how many of the instances belonging to class A were actually detected, expressed as: $\textrm{Recall} = \frac{\textrm{TP}}{\textrm{TP} + \textrm{FN}}$, where TP, FP and FN stand for true positives, false positives and false negatives. The F1 score is the harmonic mean between precision and recall, where larger values indicate better classifiers. We did not use accuracy as a metric because of the unbalanced dataset and as such would bias towards a class with higher representation and be misleading.


\subsection{Explainability of the image models}
\label{sub:explain}
To better understand the performance of the proposed model using recurrence plots, it is important to explain its decisions. One way this can be done is by using Guided Backpropagation~\cite{springenberg2015striving}, which is a way of visualising what image features CNNs detect. This is done by visualizing gradients with respect to the image where only positive values of gradients are used for backpropagating through ReLU layers, while the negative values are set to zero. Doing this, values in the filters of CNNs that are greater than zero signify pixels with higher importance to the recognition and show which pixels in the image contribute the most to the classification. 

\subsection{Resource consumption}
\label{subsec:resource}
For evaluating the resource consumption of the classifiers we consider the number of weights of the model, the FLOPs as discussed in Section \ref{subsec:considerations} and the theoretical energy consumption per prediction (TEC). To calculate TEC we used a formula provided by \cite{balaji} $TEC = \frac{FLOPs}{FLOPS/Watt}$ where $FLOPS/Watt$ represent floating point operations per second per watt. Since NVidia RTX 2080 Ti was used for conducting these experiments, its theoretical FLOPS/Watt are 53.8 GFLOPS for float32 computations were used for calculating the TEC.

\section{Results}
\label{sec:evaluation}
In this section, we evaluate the relative performance of the time series transformations considered in Section~\ref{sec:TS_transform} against a baseline and the deep learning models proposed in Section~\ref{sec:DL_models} for solving the binary and multi class classification problems formulated in Section~\ref{sec:problem_formulation}. We also provide the explanation of the model to provide transparency to how the decisions are made  and quantify relative resource consumption. The experimental methodology employed for obtain the results is detailed in Section\ref{sec:methodology}.
\subsection{Performance of image transformation models for binary classification}
\label{subsec:perform:binary}
Table~\ref{tab:results-classification} presents the results of the classifiers. The first two lines of the table list the results of the binary classifier while the last five lines list the results of the multi-class classifier. The first column of the table lists the type of classifier, the second lists the classes while the remaining four columns list the four types of considered input data and the results of the corresponding models using the three selected metrics. According to Table~\ref{tab:results-classification}, the best performing model is based on the RP as also bolded in the corresponding column. 

The reason RP model outperforms the GAF models is down to the way transformations are calculated. As explained in Section~\ref{sec:methodology}, while calculating the RP transformation, the thresholding and binarization were omitted which means that none of the information about the time series sample is lost. The images for GASF and GADF transformation are computed from the angles of polar coordinates, while disregarding the information about the amplitude, which means that some of the information about the time series sample is lost, contributing to a worse classification performance compared to RP model.

As it can be seen in the first line of the binary classifier results in Table~\ref{tab:results-classification} the RP model achieves near perfect F1 score of 0.99 in detecting anomalous links. This is slightly better than the 0.98 F1 score of GASF model, while GADF model performs the worst out of the three with the F1 score of 0.91. What can be observed is that all three image models outperform the  TS snapshot model that yields an F1 score of 0.90, which is lower by up to  0.09 compared to the F1 scores of the image based classification models.

What can also be observed from the first row is that the GADF F1 score is lower by 0.07 compared to the GASF model which shows worse performance in detecting anomalies. The reason for this is due to the ratio between the values in the region of the anomaly and the region around the anomaly which is much higher on average for GASF than for GADF, making the anomalies more prominent for the model to learn. This can best be seen in Figures~\ref{fig:example:norecovery:gaf} and ~\ref{fig:example:step-recovery:gaf} respectively.

Compared to the  TS snapshot model, significant improvement in performance is achieved in detecting non anomalous links with image models. Looking at the second line of binary classifier results in Table~\ref{tab:results-classification} it can be seen that both the RP and GADF models perform with an F1 score of 0.97, which is slightly better than the performance of GASF model with F1 score of 0.94. All three models are superior to  TS snapshot model which achieved an F1 score of  0.65.

\subsection{Performance of image transformation models for multiclass classification}
In the lower part of the Table~\ref{tab:results-classification} the last five lines list the results of the multi-class classifier, where the best performing model is based on the RP as also bolded in the corresponding column. To provide additional insights into the classification decisions, in Figure~\ref{fig:example:XAI} we provide explanation maps of a randomly extracted sample for each anomaly from the testing dataset. Lighter pixels show higher importance to the recognition, while darker pixels show lesser importance.

The general observations regarding the performance of the image based models are similar to the ones seen in Section~\ref{subsec:perform:binary}. In general GASF model is better at detecting anomalies than GADF model, while RP is the best image based model, the reasons for both observations being the same as explained in the Section~\ref{subsec:perform:binary}. Finally, according to the F1 scores all image models outperform the  TS snapshot model, where the F1 scores can be higher by up to  0.24.

As it can be seen in the first line of the multi-class classifier results in Table~\ref{tab:results-classification}, all three imaging models perform very well in predicting the SuddenD anomaly, with the RP and GADF models having an F1 score of 1.00, while the F1 score of GASF is 0.99. All three results are similar to or even slightly better than the baseline  TS snapshot model. As it can be observed in Figure~\ref{fig:example:norecovery:XAI} for detecting the SuddenD anomaly the model focuses on the black parts of the image seen in the example Figure~\ref{fig:example:norecovery:rp}. This also complies with the synthetic injection approach from  Table~\ref{tab:injection-scenario}, where SuddenD anomaly is observed towards the end of the window of the time series.

Looking at the second row of the multi-class classifier results in Table~\ref{tab:results-classification}, on average there is a slight drop in performance in SuddenR detection compared to SuddenD. The best performing model is the RP model with F1 score of 1.00, followed by GADF with an F1 score of 0.98.  Considering the F1 score of 0.85 of the  TS snapshot model, all three image based models outperform the TS snapshot model.

In the third row of the of the multi-class classifier results in Table~\ref{tab:results-classification}, the performance of the InstaD classifier can be observed. Again, with an F1 score of 0.92, RP model performs the best out of all image models, while the GASF model with an F1 score of 0.90 is a close second.  All three image based models outperform the 0.68 F1 score of the TS snapshot baseline model. Figures \ref{fig:example:recovery:XAI} and \ref{fig:example:spikes:XAI} representing the explanation maps for the SuddenD and InstaD anomalies show that for the respective anomalies the model focuses on black cross-like areas similar to the typical representations in Figures \ref{fig:example:step-recovery:rp} and \ref{fig:example:spikes:rp}. Since SuddenR and InstaD can randomly occur anywhere along the time series length, as shown in Table~\ref{tab:injection-scenario}, some activation can also be seen in other parts of the figures, to help determine the anomaly. 

According to the fourth line of the of the multi-class classifier results in Table~\ref{tab:results-classification}, compared to the 0.95 F1 score of the GASF model and 0.85 F1 score of the GADF model, the RP model is the best performing model with an F1 score of 0.99.  Both RP and GASF outperform the 0.89 F1 score of the baseline TS snapshot model while GADF underperforms. The explanation map of the final anomaly, SlowD, is presented in the Figure~\ref{fig:example:slow:XAI}. As it can be seen the most important parts of the images are along the upper and right edge. This shows that the model is most focused on the last part of the anomaly trace where the values are lower compared to the beginning of the trace, as seen in Figure~\ref{fig:example:slow:ts}, and also results in higher density area in Figure~\ref{fig:example:slow:rp} along the top and right edge of the figure.

In the last row of the of the multi-class classifier results in Table~\ref{tab:results-classification}, performance results of classifying links without anomaly are presented. Like in all of the previous rows also here the RP model with an F1 score of 0.99 is again, compared to the 0.98 F1 score of the GASF model and 0.97 F1 score of the GADF model, the best performing image model.  All three models outperform or are equal to the baseline model which achieved and F1 score of 0.97.
\begin{figure}[thb]
	\centering
	\includegraphics[width=\linewidth]{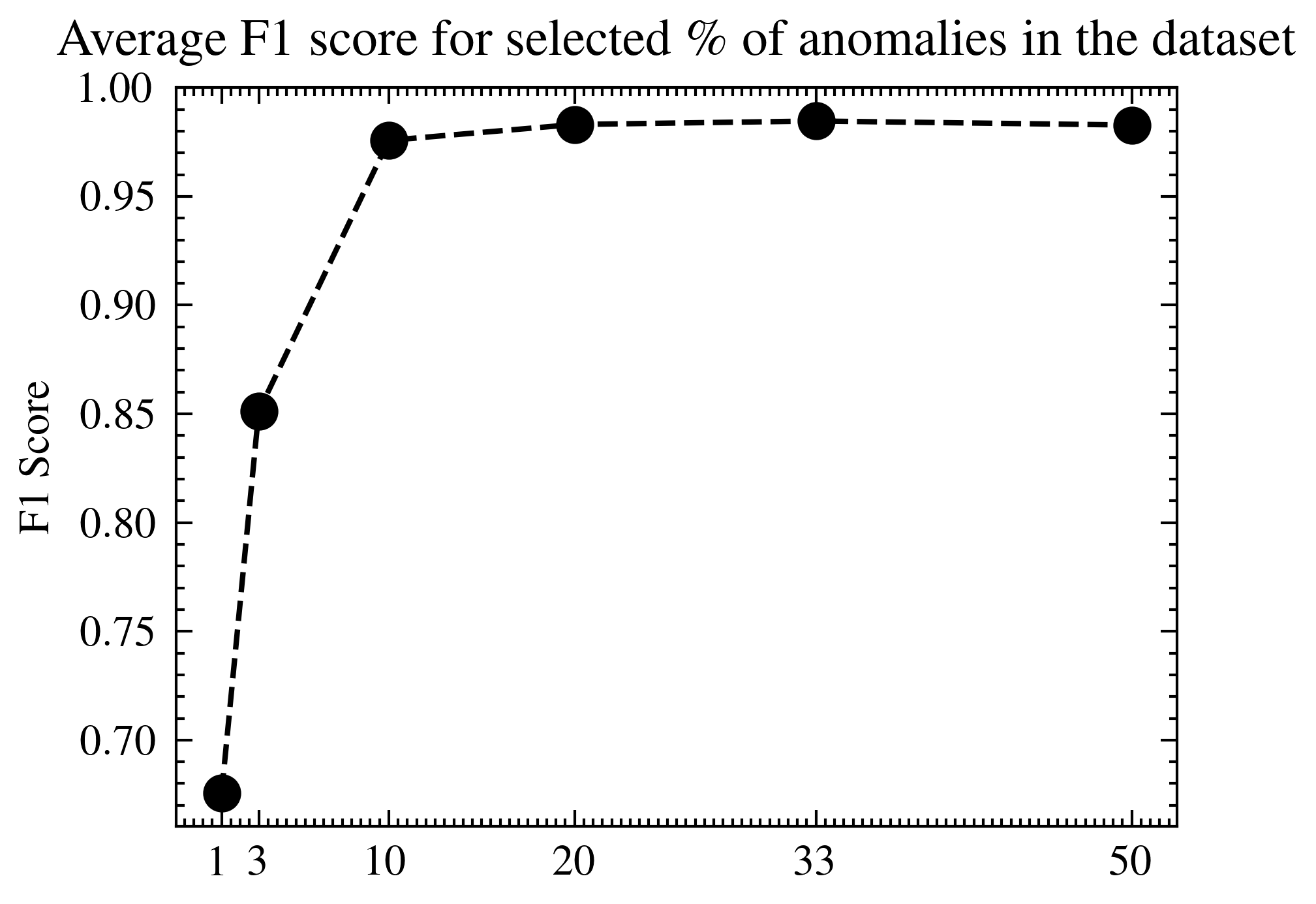}
	\caption{Average F1 score for selected percentage of anomalies in the dataset.}
	\label{fig:percentage:anomalies}
\end{figure}

\subsection{Effect of anomaly share in dataset on classifier performance}
 Since in real world settings the presence of anomalies is sparse, we provide insights into how the percentage of examples in the training dataset affects the quality of the multi-class classifier. For this we injected anomalies at the following rates 1\%, 3\%, 10\%, 20\%, 33\% and 50\% in the original Rutgers dataset, according to the parameters from Table~\ref{tab:injection-scenario}. The datasets were then transformed into RP images, since they produce the best performing model according to Table~\ref{tab:results-classification}, and fed into the proposed DL model from Section~\ref{sec:DL_models}. The results are depicted in Figure~\ref{fig:percentage:anomalies} where the x-axis represents the percentage of anomalies and the y-axis the average F1 score across 5-folds. It can be seen that a dataset with at least 10\% of injected anomalies is already sufficient for training a good enough classifier with an average F1 score of 0.975. The average F1 score for the 20\%, 33\% and 50\% ranged between 0.982 and 0.984.
\begin{table*}[tbh]
	\centering
	\ra{1.2}
	\footnotesize
	\begin{threeparttable}[b]
		\caption{Comparison to multi-class classical machine learning algorithms specialized for time series classification.}
		\label{tab:results-comparison-TSmodels}
		\begin{tabular}{lllllllllllllllll}
			\toprule

			\multirow{2}{*}{Class}
			& \multicolumn{3}{c}{TS-KNN}
			& \phantom{}
			& \multicolumn{3}{c}{TS-SVM}
			& \phantom{}
			& \multicolumn{3}{c}{Our model}
			\\\cmidrule{2-4}\cmidrule{6-8}\cmidrule{10-12}
			
			& Prec.
			& Rec.
			& F1
			& 
			& Prec.
			& Rec.
			& F1
			& 
			& Prec.
			& Rec.
			& F1
			\\\midrule
			
			
			SuddenD	& 1.00 & 1.00 &  \textbf{1.00} &	& 1.00 & 1.00 &  \textbf{1.00} &	& 1.00 & 1.00 &  \textbf{1.00}\\
			
			SuddenR	& 0.81 & 0.83 & 0.82 &	& 0.98 & 0.92 & 0.95 &	& 1.00 & 1.00 &  \textbf{1.00}\\
			
			InstaD	& 0.78 & 0.61 & 0.68 &	& 0.94 & 0.80 & 0.86 &	& 0.93 & 0.90 &  \textbf{0.92}\\
			
			SlowD	& 0.87 & 0.46 & 0.60 &	& 0.93 & 0.87 & 0.90&	& 1.00 & 0.99 &  \textbf{0.99}\\
			
			No anomaly	& 0.91 & 0.98 & 0.94 &	& 0.96 & 0.99 & 0.97&	& 0.99 & 0.99 &  \textbf{0.99}\\
			\bottomrule
		\end{tabular}	
	\end{threeparttable}
\end{table*}
\begin{table*}[!tbp]
	\centering
	\ra{1.2}
	\footnotesize
	\begin{threeparttable}[b]
		\caption{Comparison of well known DL models (AlexNet and VGG11.)}
		\label{tab:results-comparison-DLModels}
		\begin{tabular}{lllllllllllllllll}
			\toprule

			\multirow{2}{*}{Class}
			& \multicolumn{3}{c}{AlexNet \cite{krizhevsky2012imagenet}}
			& \phantom{}
			& \multicolumn{3}{c}{VGG11 \cite{simonyan2014very}}
			& \phantom{}
			& \multicolumn{3}{c}{Our model}
			\\\cmidrule{2-4}\cmidrule{6-8}\cmidrule{10-12}
			
			& Prec.
			& Rec.
			& F1
			& 
			& Prec.
			& Rec.
			& F1
			& 
			& Prec.
			& Rec.
			& F1
			\\\midrule
			
			
			SuddenD	& 0.99 & 1.00 & 0.99 &	& 1.00 & 1.00 &  \textbf{1.00} &	& 1.00 & 1.00 &  \textbf{1.00}\\
			
			SuddenR	& 0.99 & 0.99 & 0.99 &	& 1.00 & 1.00 &  \textbf{1.00}&	& 1.00 & 1.00 &  \textbf{1.00}\\
			
			InstaD	& 0.74 & 0.92 & 0.82 &	& 0.90 & 0.92 & 0.91&	& 0.93 & 0.90 &  \textbf{0.92}\\
			
			SlowD	& 0.93 & 0.95 & 0.94 &	& 0.97 & 1.00 &  \textbf{0.99}&	& 1.00 & 0.99 &  \textbf{0.99}\\
			
			No anomaly	& 0.99 & 0.97 & 0.98 &	& 0.99 & 0.99 &  \textbf{0.99}&	& 0.99 & 0.99 &  \textbf{0.99}\\
			\midrule
			Number of weights & & $\approx$80M	&&&& $\approx$190M 	&&&& $\approx$6M\\
			\midrule
			Number of GFLOPs & &$\approx$4	&&&& $\approx$26.5	&&&& $\approx$2.1\\
			\midrule
			TEC (Joules) & &$\approx$0.07	&&&& $\approx$0.49	&&&& $\approx$0.04\\
			\bottomrule
		\end{tabular}	
	\end{threeparttable}
\end{table*}

\subsection{Comparison to multi-class classical machine learning algorithms specialized for time series classification}	
 The results are presented in Table~\ref{tab:results-comparison-TSmodels} where the first column lists the classes while the remaining two columns list the two best performing TS classification specialized ML models and the proposed RP model using the three selected metrics.  As it can be observed from the table both TS k-Nearest neighbor model (TS-KNN) and TS Support vector machine model (TS-SVM) are capable of correctly classifying the SuddenD anomaly with perfect F1 score of 1.00 same as our model. For detecting the remaining classes our model outperforms the other two with an up to 39 percentage point difference in the F1 score compared to TS-KNN model and up to 9 percentage points in the F1 score compared to TS-SVM. This shows that the proposed model using TS image transformation can outperform classical ML models that are specialized for TS classification in multi-class recognition of wireless anomalies.
\begin{table}[htbp]
	\centering
	\ra{1.2}
	\footnotesize
	\begin{threeparttable}[b]
		\caption{Comparison of results between our model and the state of the art in \cite{cerar2020anomaly}}
		\label{tab:results-comparison-gregor}
		\begin{tabular}{lllllllllllllllll}
			\toprule

			\multirow{2}{*}{Class}
			& \multicolumn{3}{c}{Ensemble learner from \cite{cerar2020anomaly}}
			& \phantom{}
			& \multicolumn{3}{c}{Our best result}
			\\\cmidrule{2-4}\cmidrule{6-8}
			
			& Prec.
			& Rec.
			& F1
			& 
			& Prec.
			& Rec.
			& F1
			\\\midrule
			
			
			SuddenD	& 1.00 & 1.00 &  \textbf{1.00} &	& 1.00 & 1.00 &  \textbf{1.00}\\
			
			SuddenR	& 0.49 & 0.67 & 0.56 &	& 1.00 & 1.00 &  \textbf{1.00}\\
			
			InstaD	& 0.31 & 0.42 & 0.36 &	& 0.93 & 0.90 &  \textbf{0.92}\\
			
			SlowD	& 0.54 & 0.81 & 0.65&	& 1.00 & 0.99 &  \textbf{0.99}\\
			
			No anomaly	& 0.93 & 0.87 & 0.90 &	& 0.99 & 0.99 &  \textbf{0.99}\\
			\bottomrule
		\end{tabular}	
	\end{threeparttable}
\end{table}
\begin{table*}[!tbh]
	\centering
	\ra{1.2}
	\normalsize
	\scalebox{0.8}{
	\begin{threeparttable}[b]
		\caption{Comparison of results between our model and the state of the art in \cite{bertalanic_deep_nodate}}
		\label{tab:results-classification-wimob}
		
		\begin{tabular}{llllllllllllllllllll}
			\toprule
			
			\multirow{2}{*}{Class}
			& \multicolumn{3}{c}{DLC}
			& \phantom{}
			& \multicolumn{3}{c}{LRC}
			& \phantom{}
			& \multicolumn{3}{c}{RFC}
			& \phantom{}
			& \multicolumn{3}{c}{SVMC}
			& \phantom{}
			& \multicolumn{3}{c}{Our best result}
			\\\cmidrule{2-4}\cmidrule{6-8}\cmidrule{10-12} \cmidrule{14-16} \cmidrule{18-20}
			
			& Prec.
			& Rec.
			& F1
			& 
			& Prec.
			& Rec.
			& F1
			& 
			& Prec.
			& Rec.
			& F1
			& 
			& Prec.
			& Rec.
			& F1
			& 
			& Prec.
			& Rec.
			& F1
			\\\midrule

			SuddenD	& 1.00 & 1.00 &  \textbf{1.00} &	& 1.00  & 1.00 &  \textbf{1.00} &	& 1.00 & 1.00 &  \textbf{1.00} &	& 1.00 & 1.00 &  \textbf{1.00} &	& 1.00 & 1.00 &  \textbf{1.00} \\
			
			SuddenR	& 0.98 & 0.96 & 0.97 &	& 0.49 & 0.67 & 0.57 &	& 0.99 & 0.77 & 0.87 &	& 0.76 & 0.68 & 0.72 & & 1.00 & 1.00 &  \textbf{1.00} \\
			
			InstaD	& 0.97 & 0.81 & 0.88 &	& 0.10 & 0.20 & 0.13 &	& 0.08 & 0.01 & 0.02 &	& 0.37 & 0.07 & 0.11 &	& 0.93 & 0.90 &  \textbf{0.92} \\
			
			SlowD	& 0.62 & 0.85 & 0.72 &	& 0.54 & 0.81 & 0.65 &	& 1.00 & 0.21 & 0.34 &	& 0.37 & 0.68 & 0.47 &	& 1.00 & 0.99 &  \textbf{0.99} \\
			
			No anomaly	& 0.97 & 0.95 & 0.96 &	& 0.93 & 0.77 & 0.84 &	& 0.85 & 0.99 & 0.92 &	& 0.88 & 0.89 & 0.89 &	& 0.99 & 0.99 &  \textbf{0.99} \\
			\bottomrule
		\end{tabular}
	\end{threeparttable}}
\end{table*}
\subsection{Comparison to well known DL models}
  The results are presented in the Table~\ref{tab:results-comparison-DLModels} where the first column lists the classes while the remaining two columns lists both selected DL models and our best performing RP model using the three selected metrics. The first five lines represent the per anomaly type performance while the last three quantify the models' resource consumption through the number of weights, required floating point operations (FLOPs) and theoretical energy consumption per prediction (TEC) as detailed in Section \ref{subsec:resource}.

 The general observation regarding the performance is that our model with, 6 million trainable weights, works on par with VGG11 consisting of 190 million weights, while it outperforms the AlexNet model with 80 million weights in terms of F1 score. From the first two rows of the table it can be seen that according to F1 score all three models had a very similar performances. For classification of SuddenD and SuddenR anomalies ALexNet achieved an F1 score of 0.99 for both classes, while both VGG11 and our model performed with the perfect F1 score of 1.00. The main difference between the models was in classification of InstaD and SlowD anomalies, where for both AlexNet scores the lowest F1 score out of all three models. On the other hand VGG11 and our model perform very similarly with our model slightly outperforming VGG11 in classification of InstaD anomaly. These results show that our proposed DL model can work better than some of the well know DL architectures, while having up to 97\% less weights and needs up to 13 times less FLOPs, while the TEC per prediction for all three models is negligible.
\subsection{Comparison to the state of the art in the wireless link layer anomaly detection}
 The results for the comparison can be seen in Table~\ref{tab:results-comparison-gregor} and Table~\ref{tab:results-classification-wimob}. In both tables the first column lists the classes while the remaining columns lists classifiers from \cite{cerar2020anomaly} or \cite{bertalanic_deep_nodate} and our best performing RP model using the three selected metrics.

 From Table~\ref{tab:results-comparison-gregor} it can be seen that with the exception of the SuddenD anomaly detection, our model outperforms their ensemble learner in all other classes. The paper showed the potential of such models but this result shows that those trained models are not yet ready for use in production because they have trouble performing when put into an ensemble. They made assumptions that only isolated cases of various anomalies can appear on links, which is not entirely aligned with a realistic environment where various anomalies can appear simultaneously. Due to these assumptions, the classifiers put into an ensemble can classify anomalies, that are not their own, as false positive. This goes to show that our model is more robust when it comes to an unseen mix of anomalies, compared to the model from \cite{cerar2020anomaly}.

 The classifiers trained in \cite{bertalanic_deep_nodate} used raw time series data as an input and the authors used extensive grid search of optimal parameters in designing their models. It can be seen that with the exception of SuddenD anomaly, our model outperforms their Deep Learning Classifier (DLC), Logistic Regression Classifier (LRC), Random Forest Classifier (RFC) and Support Vector Machine Classifier (SVMC) on every anomaly class. In terms of performance the DLC is the closest to our model. In terms of F1 score the classification of SuddenR, SlowD and No Anomaly classes is worse by only up to 0.04, while the F1 score for the classification of SlowD is worse by 0.27 compared to our model. These results additionally show the potential of our proposed approach for the classification of wireless anomalies.

\section{Conclusions}
\label{sec:conclusions}
In this paper, we performed a first time analysis of image-based representation techniques for wireless anomaly detection using recurrence plots and Gramian angular fields for binary and multi-class classification and proposed a new deep neural network architecture that is able to distinguish between the considered wireless link layer anomalies.  We also elaborated on the design considerations for the proposed resource aware model. We evaluated the proposed model from four different perspectives and concluded the following.

First, our results show that the best performing model  developed using the recurrence plot transformation outperforms a) the Gramian angular fields image-based models by up to 14 percentage points, and b) the baseline time-series snapshot image based model by up to 32 percentage points for binary classification and by up to 24 percentage points for multiclass classification. We also show how decisions taken by the model can be explained and that a training dataset with 10\% injected anomalies is already sufficient to train a good classifier. 

 Second, we show that compared to classical ML models for multiclass classification using the well-known dynamic time warping method, the proposed resource aware model that used the recurrence plot transformation   improves performance by up to 24 percentage points. 
	
 Third, we demonstrate that compared to more well known DL architectures such as AlexNet and VGG11, the proposed model outperforms or performs on par while having <10 times their weights and up to $\approx$8\% of their computational complexity. 
	
 Finally, compared to the state of the art in the application area that relies on classical machine learning based ensemble model, raw TS DL model, and three other well known classical ML algorithms, the proposed  model is more robust in detecting unseen anomalies outperforming the ensemble by up to 55 percentage point, while also outperforming raw TS DL model and other ML algorithms by up to 27 and 90 percentage points, respectively.



\section*{Acknowledgements}
The authors would like to thank dr. Gregor Cerar for his technical support and insightful discussions on the topic. Special acknowledgements go to the anonymous reviewers that provided constructive and insightful comments that significantly increased the quality of this paper.  This work was funded by the Slovenian Research agency under grants P2-0016 and PR-10467.

\ifCLASSOPTIONcaptionsoff
  \newpage
\fi
\bibliographystyle{IEEEtran}
\bibliography{IEEEabrv,IEEEexample}
\end{document}